\documentclass[10pt,journal,compsoc]{IEEEtran}

\ifCLASSOPTIONcompsoc
  \usepackage[nocompress]{cite}
\else
  \usepackage{cite}
\fi

\ifCLASSINFOpdf
\else
\fi

\usepackage[utf8]{inputenc} 
\usepackage[T1]{fontenc}    
\usepackage{hyperref}       
\usepackage{url}            
\usepackage{booktabs}       
\usepackage{amsfonts}       
\usepackage{nicefrac}       
\usepackage{microtype}      
\usepackage{xcolor}         
\usepackage{graphicx}
\usepackage{subfigure}
\usepackage{amsmath,amssymb} 
\usepackage{caption} 
\usepackage{hyperref}
\usepackage{wrapfig}
\usepackage[american]{babel}

\usepackage{hyperref}       
\usepackage{graphicx} 
\usepackage{algorithm}
\usepackage{algorithmic}
\usepackage{times}
\usepackage{soul}
\usepackage{url}
\usepackage[utf8]{inputenc}
\usepackage{amsmath}
\usepackage{amsthm}
\usepackage{booktabs}
\usepackage{algorithm}
\usepackage{algorithmic}
\usepackage{subfigure}
\usepackage{amsmath}
\usepackage{amssymb}
\usepackage{booktabs}
\usepackage{multirow}
\usepackage{paralist,algorithmic,algorithm}
\usepackage[american]{babel}
\usepackage{microtype}
\usepackage{lipsum}
\usepackage{bm}
\usepackage[switch]{lineno}  %
 
\newcommand{\x}{{\bf x}}
\newcommand{\w}{{\bm w}}
\newcommand{\p}{{\bm p}}

\newcommand{\D}{\mathcal{D}}

\newcommand{\Sup}{\mathcal{S}}
\newcommand{\Q}{\mathcal{Q}}
\newcommand{\R}{\mathbb{R}}

\newcommand{\eg}{\emph{e.g.}}
\newcommand{\ie}{\emph{i.e.}}

\newcommand{\experimentsection}[1]{\subsection{#1}}
\newcommand{\bfname}[1]{{\bf #1}}
\newcommand{\name}{{\sc Limit }}
\newcommand{\mame}{{\sc Limit}}

\begin{document}
%
\title{Few-Shot Class-Incremental Learning by Sampling Multi-Phase Tasks}
%
%
%
%

\author{Da-Wei Zhou, Han-Jia Ye, Liang Ma, Di Xie, Shiliang Pu,  De-Chuan Zhan
\IEEEcompsocitemizethanks{
	\IEEEcompsocthanksitem D.-W. Zhou, H.-J. Ye and D.-C. Zhan are with State Key Laboratory for Novel Software Technology, Nanjing University,  Nanjing, 210023, China; 
	Han-Jia Ye is the corresponding author. \\
	E-mail: \{zhoudw, yehj, zhandc\}@lamda.nju.edu.cn

	L. Ma, D. Xie and S. Pu are with Hikvision Research Institute, Hangzhou 310051, China. E-mail:
	\{maliang6, xiedi, pushiliang.hri\}@hikvision.com
}
}

%
%

\markboth{Journal of \LaTeX\ Class Files,~Vol.~14, No.~8, August~2015}%
{Shell \MakeLowercase{\textit{et al.}}: Bare Demo of IEEEtran.cls for Computer Society Journals}
%



\IEEEtitleabstractindextext{%
\begin{abstract}
New classes arise frequently in our ever-changing world, \eg, emerging topics in social media and new types of products in e-commerce. A model should recognize new classes and meanwhile maintain discriminability over old classes. Under severe circumstances, only \emph{limited} novel instances are available to incrementally update the model. The task of recognizing few-shot new classes without forgetting old classes is called few-shot class-incremental learning (FSCIL). In this work, we propose a new paradigm for FSCIL based on meta-learning by LearnIng Multi-phase Incremental Tasks (\mame), which synthesizes fake FSCIL tasks from the base dataset. The data format of fake tasks is consistent with the `real' incremental tasks, and we can build a generalizable feature space for the unseen tasks through meta-learning. Besides, \name also constructs a calibration module based on transformer, which calibrates the old class classifiers and new class prototypes into the same scale and fills in the semantic gap. The calibration module also adaptively contextualizes the instance-specific embedding with a set-to-set function. \name efficiently adapts to new classes and meanwhile resists forgetting over old classes. Experiments on three benchmark datasets (CIFAR100, \textit{mini}ImageNet, and CUB200) and large-scale dataset, \ie, ImageNet ILSVRC2012 validate that \name achieves state-of-the-art performance. 
\end{abstract}
}

\maketitle

\IEEEdisplaynontitleabstractindextext

%
\IEEEpeerreviewmaketitle



\section{Introduction}
\IEEEPARstart{W}{ith} the rapid development of supervised learning, current learning systems have achieved or even surpassed human-level performances in many tasks~\cite{deng2009imagenet,danelljan2014accurate,babenko2010robust,khan2008tracking,joseph2020incremental,khan2021transformers}. However, real-world applications often face the stream format data~\cite{gomes2017survey} and novel classes~\cite{zhou2016learnware,zhou2021learning,yang2021generalized,yang2021semantically}, \eg, incoming hot topics in social media and new types of products in e-commerce.
 Under such circumstances, an ideal model should recognize new classes and meanwhile maintain discriminability over old classes, which is called Class-Incremental Learning (CIL). The most challenging problem in CIL is catastrophic forgetting --- when optimizing the model with new classes, the formerly acquired knowledge on old classes is quickly forgotten~\cite{goodfellow2014empirical}. The trade-off between learning new classes and retaining old classes is called the \emph{stability-plasticity dilemma}~\cite{mermillod2013stability}. There are extensive efforts in resisting forgetting from different perspectives, \eg, knowledge distillation~\cite{li2017learning,rebuffi2017icarl}, parameter consolidation~\cite{kirkpatrick2017overcoming,aljundi2018memory} and post-tuning~\cite{zhao2020maintaining,yu2020semantic}.
 
\begin{figure}[t]
	\begin{center}
		\includegraphics[width=1\columnwidth]{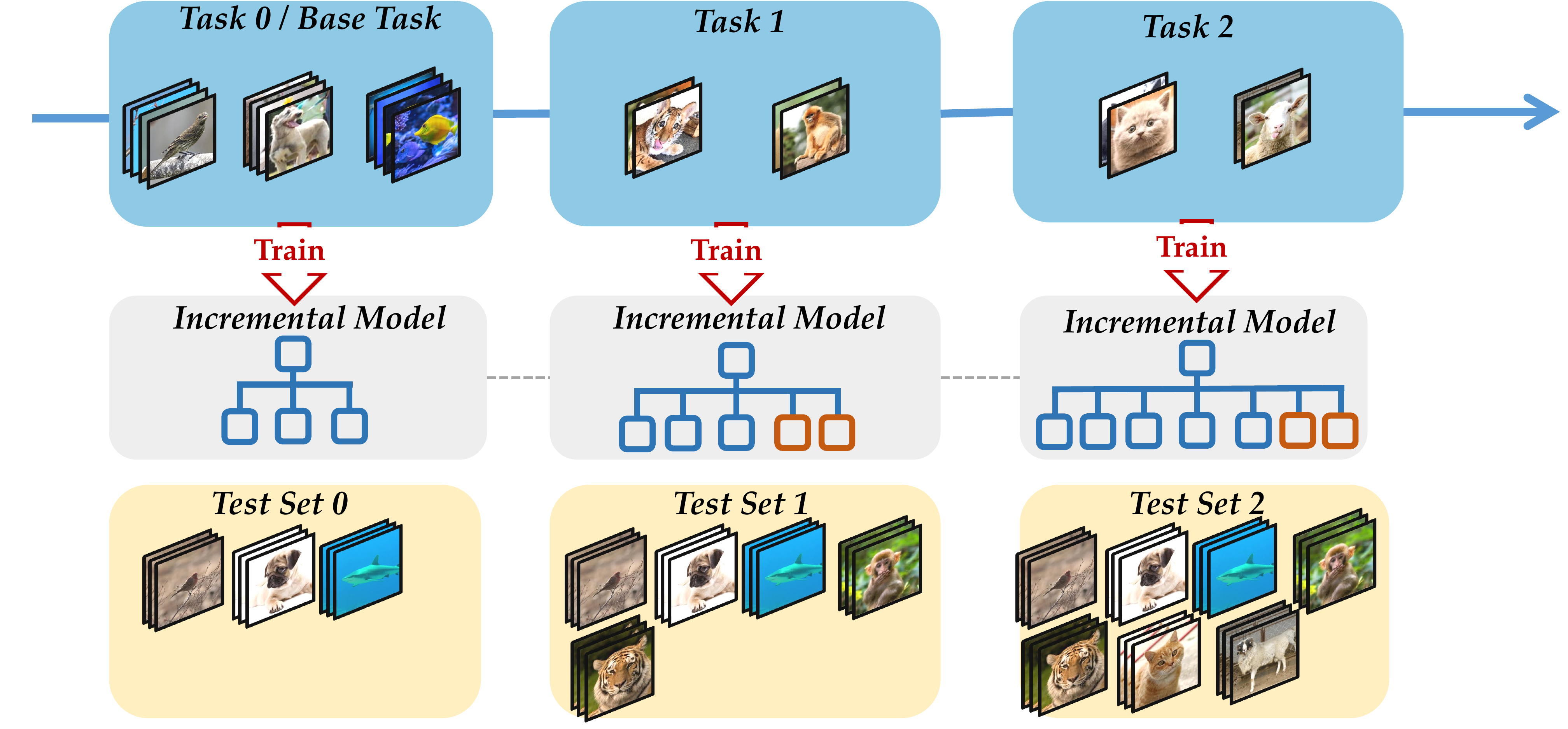}
	\end{center}
	\caption{The setting of few-shot class-incremental learning. 
		Non-overlapping classes arrive sequentially, and we need to build a classifier for all the classes incrementally. There are sufficient training images in the base task and limited images in the following tasks. 
		After the learning process of each task, the model is evaluated among all seen classes. An ideal model should perform well in the newly learned classes and remember the former without forgetting.
	} \label{figure:intro}
\end{figure}

The aforementioned methods are capable of solving the CIL problems with abundant training examples. However, the instance labeling and collection cost are sometimes unbearable in real-world applications. For example, when training an incremental model for face recognition, users are unwilling to upload copious images~\cite{tan2006face}; when training a model to classify rare birds, their images are challenging to collect.
When we need to build an incremental model for these tasks, only \emph{limited} training instances are available.
 Consequently, designing algorithms to handle Few-Shot Class-Incremental Learning (FSCIL) becomes essential for a real-world classification system. 
The setting of FSCIL is shown in Figure~\ref{figure:intro}. There are sufficient training instances in the base session to build the initial model, which needs to continually incorporate new classes with only limited data points. The learning process should not harm the former knowledge since the model is evaluated over all seen classes.
 Similar to CIL, learning new classes will cause catastrophic forgetting of former ones. Additionally, since new class instances are insufficient, it is easy to observe the \emph{overfitting} phenomena on  these limited inputs, which increases the learning difficulty of incremental tasks.

An intuitive way for few-shot class-incremental learning is to directly adopt CIL methods, which get poor performance~\cite{tao2020few,zhang2021few}. The problem is caused by the limited instances --- training few-shot instances will trigger the overfitting phenomenon and destroy the pre-trained feature embeddings~\cite{finn2017model}. As a result, the destroyed embedding is unable to capture the characteristics of former classes, which further exacerbates catastrophic forgetting.  
Correspondingly, there are several works considering extending the model with limited instances and resisting overfitting~\cite{zhang2021few,zhu2021self}. From the few-shot learning perspective, they seek to synthesize the incremental learning scenario and make the model capable of extending in a single stage. However, FSCIL is a \emph{long-term} learning process --- we should cultivate the long-term learning ability of the model to make it consistent with real scenarios.

Accordingly, we propose a new training paradigm to consider multi-phase training for FSCIL. We would like to equip the model with the \emph{learning ability} of FSCIL tasks --- if the model can tackle different kinds of simulated fake-FSCIL tasks, it will easily handle the incoming `real' FSCIL task with generalizability.
Solving these fake tasks enables the model to simultaneously differentiate known classes and multiple new classes. 
 Take the learning process of a child for an example. Suppose he is trained to recognize different kinds of few-shot animals incrementally. In that case, he shall obtain the ability of incremental learning and can recognize few-shot vehicles incrementally, which is as easy as pie.

On the other hand, since the number of training instances for old and new classes differs,
there is an inherent \emph{information gap} between the many-shot old classes and few-shot new classes. The model is well-optimized and good at depicting the old classes' features but ill-optimized for new classes. As a result, directly aggregating the prediction results between them will result in biased predictions.
 To this end, a calibration process should be conducted to derive a balance between old and new classes and re-rank the output probabilities among all seen classes. A suitable calibration module can reflect the context information between query instances and classifiers, and co-adapt them to remove the bias of the incremental model.

In this paper, we propose LearnIng Multi-phase Incremental Tasks (\mame) for few-shot class-incremental learning to learn a more generalizable feature embedding. In detail, we sample vast fake-FSCIL tasks from the base dataset and enable the model to handle incoming FSCIL tasks. By optimizing such fake-tasks, we extract invariant information between the base and incremental sessions,\footnote{We use `session,' `phase,' `task' interchangeably in this paper.} and prepare the model for incoming few-shot incremental sessions.
Additionally, we propose encoding the meta-learned knowledge into the meta-calibration module, which aims to learn the calibration relationship between few-shot prototypes and many-shot old class classifiers.
The meta-calibration module is implemented with transformers to capture the long dependencies and adapt the embeddings with discriminative features. 
 With the optimized meta-calibration module, \name is capable of calibrating the base classifiers with the incoming incremental sessions. \name maintains the discriminability of old classes when learning new classes with humble forgetting.
Various experiments against state-of-the-art methods validate the superiority of \name not only on benchmark few-shot class-incremental learning datasets but also on the large-scale datasets, \ie, ImageNet ILSVRC2012~\cite{deng2009imagenet}. Visualization effects on decision boundary and output probabilities demonstrate the effectiveness of training fake-incremental tasks and the meta-calibration module.

Our contributions can be summarized as follows:
\begin{itemize}
	\item We propose a new learning paradigm by sampling multi-phase fake-incremental tasks to prepare the model for incoming updates in few-shot class-incremental learning;
	\item We utilize the transformer architecture as the set-to-set function, which encodes the inductive bias and produces instance-specific embedding for better calibration;
	\item Benefiting from the learning ability of \name to adapt to new tasks, \name achieves state-of-the-art performance on three benchmark datasets.
\end{itemize}

The rest of this paper is organized as follows. We discuss some related work about few-shot class-incremental learning in Section 2 and introduce the preliminaries in Section 3. The proposed \name is presented in Section 4. Afterward, we present our experimental results and discussions in Section 5. Section 6 concludes the paper with future issues.

\section{Related Work}
The few-shot class-incremental learning problem is related to several topics, and we introduce them separately.
  
\subsection{Class-Incremental Learning (CIL)} Class-Incremental Learning is now a popular topic in the machine learning field~\cite{wang2022foster,de2019continual,masana2020class,tao2020topology,zhou2021pycil}, which can be roughly divided into two groups by whether saving exemplars of old class instances.  
Non-exemplar-based methods consider using regularization terms to consolidate model output or dynamic model structures to meet the requirement of new classes. EWC~\cite{kirkpatrick2017overcoming} measures the importance of each parameter with Fisher information matrix and expects the important parameters to be changed slightly with regularization terms. Apart from Fisher information matrix, there are other ways to measure the parameter importance, \ie, MAS~\cite{aljundi2018memory} and SI~\cite{zenke2017continual}. 
Other non-exemplar-based methods change the network structure by expanding and pruning to meet the requirements of new tasks~\cite{xu2018reinforced,yoon2018lifelong,yan2021dynamically,zhou2022model}. 

Exemplar-based methods save representative instances from each class and rehearsal them when learning new classes. iCaRL~\cite{rebuffi2017icarl} replays and conducts knowledge distillation to maintain former knowledge. BiC~\cite{wu2019large} utilizes these exemplars to build a validation set and optimizes an extra rescale layer. GEM~\cite{chaudhry2018efficient,lopez2017gradient} uses exemplars for gradient projection to overcome forgetting. Other works consider saving embeddings instead of raw images~\cite{iscen2020memory}, using generative models for data rehearsal~\cite{xiang2019incremental}, task-wise adaptation~\cite{rajasegaran2020itaml}, and output normalization~\cite{zhao2020maintaining,belouadah2019il2m,hou2019learning}.

\subsection{Few-Shot Learning (FSL)}
Few-Shot Learning aims to extract the inductive bias from base classes and generalize it to unseen classes~\cite{wang2020generalizing,chen2018closer,ye2022few,ye2021train}.  Current FSL methods can also be roughly divided into two groups: optimization-based methods and metric-based methods. 
Deep models learn through backpropagation of gradients, while backpropagation cannot converge within a small number of optimization steps. To this end, optimization-based methods try to enable fast model adaptation with few-shot data~\cite{RaviL17}. MAML~\cite{finn2017model,nichol2018first} is a general optimization algorithm compatible with other models that learn through gradient descent, which learns a common model initialization for few-shot tasks. MAML is widely applied into vast learning scenarios, \eg, robotics control~\cite{clavera2018learning}, neural translation~\cite{gu2020meta} and uncertainty estimation~\cite{finn2018probabilistic}, and many improved algorithms~\cite{arnold2021maml,antoniou2018train,deleu2018effects,lee2018gradient,ye2020few} have been made to further boost its performance.

On the other hand, metric-based methods consider learning suitable distance metrics between support and query instances.  Siamese Network~\cite{koch2015siamese} uses $\ell_1$ distance. Matching Network~\cite{vinyals2016matching} adopts the cosine similarity. Relation Network~\cite{sung2018learning} trains a learnable distance metric with a neural network. Prototypical Network~\cite{snell2017prototypical} utilizes the Euclidean distance between the class center and query instance. DeepEMD~\cite{zhang2020deepemd} applies the Earth Mover's Distance.
Apart from the classification problem, FSL also has a close relationship with other computer vision tasks, \eg, object detection~\cite{xu2017few}, semantic learning~\cite{yin2021hierarchical}, video classification~\cite{cao2020few} and face alignment~\cite{browatzki20203fabrec}.

\subsection{Few-Shot Class-Incremental Learning}
Few-shot class-incremental learning is recently proposed to tackle few-shot inputs in the class-incremental setting~\cite{zhou2022forward,zhao2021mgsvf}. TOPIC~\cite{tao2020few} defines the benchmark-setting of FSCIL, which proposes the neural gas structure to preserve the topology of features between old and new classes.  Exemplar Relation Graph~\cite{DongHTCWG21} maintains a graph to represent the class-wise relationship, which is utilized for knowledge distillation.
FSLL~\cite{mazumder2021few} adopts the idea from CIL algorithms, which selects a few parameters to be updated at every incremental session to resist overfitting.
Noticing the characteristics of few-shot instances, semantic-aware knowledge distillation~\cite{Cheraghian_2021_CVPR} considers using
auxiliary word embedding of few-shot instances to boost model updating. Self-promoted prototype refinement~\cite{zhu2021self} considers the \emph{one-stage} extension ability and adapts the feature representation to various generated incremental episodes.
The current state-of-the-art method is CEC~\cite{zhang2021few}, which decouples the training process into embedding learning and classifier learning. It also adopts an extra graph model to propagate context information between classifiers for adaptation. 
However, ~\cite{zhang2021few,zhu2021self} only consider extending the model in a single phase, while FSCIL is a dynamic process with multiple phases. By contrast, our \name considers cultivating the extending ability with multiple phases. 

There is a similar setting proposed by~\cite{qi2018low,gidaris2018dynamic,yoon2020xtarnet}, namely Generalized Few-Shot Learning (GFSL). It also addresses the scenario where a pre-trained model is going to learn new classes with limited instances. The target of GFSL is to maintain the classification performance over both old and new classes.
However, there is only a single incremental stage in GFSL, which is less challenging than FSCIL. Typical GFSL algorithms try to solve the problem by classifier weight generalization~\cite{gidaris2018dynamic,ye2021learning}, subspace regularization~\cite{akyurek2021subspace} and attention-based regularization~\cite{ren2019incremental}.

\subsection{Transformer Models in Vision}
Transformer models~\cite{vaswani2017attention,khan2021transformers,lin2017structured} are first proposed to handle natural language processing tasks, \eg, text classification~\cite{kenton2019bert,BrownMRSKDNSSAA20}, question answering~\cite{zhao2020condition} and machine translation~\cite{wang2019learning}. Inspired by the breakthroughs in natural language processing filed, researches in the computer vision community have adapted them for CV tasks, \eg, sequence transformer~\cite{chen2020generative},vision transformer~\cite{han2020survey,wang2021pyramid,dosovitskiy2020image}. Transformer models are now widely applied into visual question answering~\cite{zhou2021trar}, object detection~\cite{carion2020end}, segmentation~\cite{liang2020polytransform}, video understanding~\cite{girdhar2019video}, etc. In this paper, we adopt transformer as the set function, which helps to encode the calibration information between many-shot old classes and few-shot new classes.

\begin{figure*}[t]
	\begin{center}
		\includegraphics[width=1.95\columnwidth]{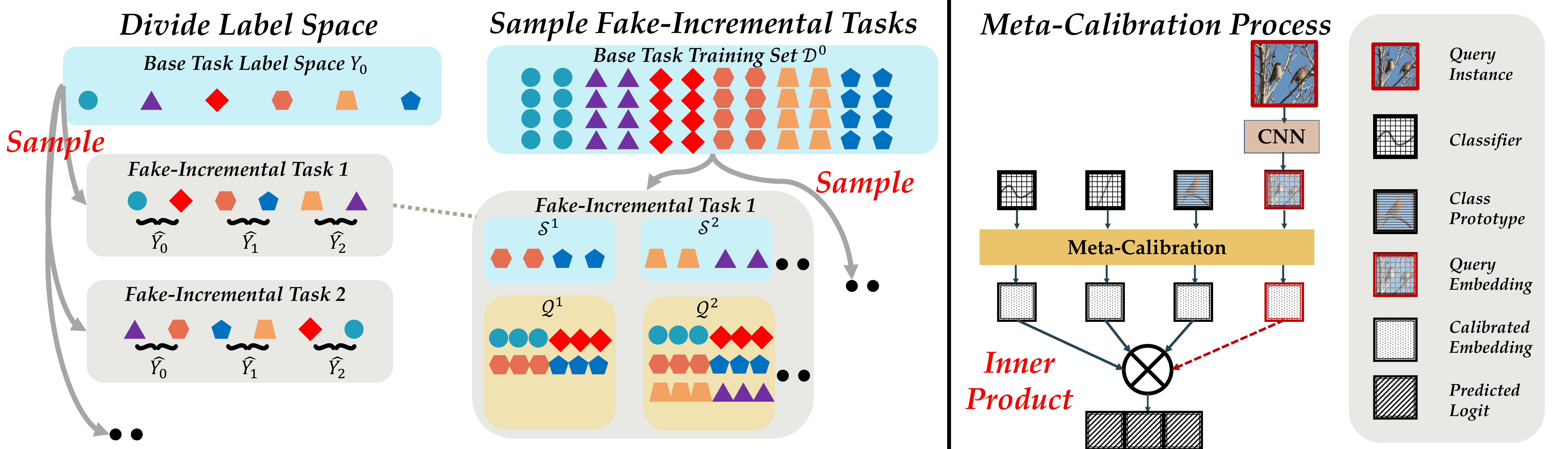}
	\end{center}
	\caption{Illustration of \mame. Left: We sample fake-incremental tasks from the base training set $\D^0$, forming various fake-tasks.  
		Right: Meta-calibration process. The model needs to calibrate between old class classifiers and new class prototypes with a set-to-set function. We also input the query instance embedding into the meta-calibration module to contextualize the classification task, generating instance-specific embeddings. The final logit is calculated by the inner-product of the transformed classifier and transformed query embedding.
	} \label{figure:teaser}
\end{figure*}

\section{From Old Classes to New Classes}
In this section, we first introduce the setting of FSCIL. Afterward, we show two approaches and discuss their limitations for FSCIL.
\subsection{Few-Shot Class-Incremental Learning}
Under the static environment, a  model receives the training set with \emph{sufficient} instances: $\D^{0}=\left\{\left(\x_{i}, y_{i}\right)\right\}_{i=1}^{n_0}$, which contains \emph{i.i.d.} samples from the distribution $\D^{0}_t$. $\x_i \in \R^D$ is a training instance from class $y_i \in Y_0$ and $Y_0$ is the corresponding label space. An algorithm fits a model $f(\x)$ to minimize the expected risk over the instance distribution $\D^{0}_t$:
\begin{align} \label{eq:closed}
	 \mathbb{E}_{\left(\x_{j}, {y}_{j}\right) \sim \D_t^{0}}\left[\ell\left(f
	 \left(\x_{j} 
	 \right), {y}_{j}\right)\right] \,,
\end{align}
where $\ell(\cdot,\cdot)$ measures the discrepancy between prediction and ground-truth label. 
The classification model can be decoupled into two parts: embedding  function $\phi(\cdot):\mathbb{R}^{D} \rightarrow \mathbb{R}^{d}$ and linear classifier $W\in\mathbb{R}^{d\times |{Y}_{0}|}$, \ie, $f(\x)=W^{\top}\phi(\x)$. We denote the classifier for class $k$ as $\w_k$, \ie, $W=[\w_1,\cdots,\w_{|Y_0|}]$. 
In FSCIL, the first task, \ie,  $\D^0$, usually contains a \emph{sufficient} amount of data, which is called the base task/session.

\noindent \textbf{Few-shot class-incremental learning:} In our dynamic world, training sets often arrive incrementally with \emph{limited} instances, \ie, a sequence of training datasets $\left\{\D^{1}, \cdots, \D^{B}\right\}$ will emerge sequentially. $\D^{b}=\left\{\left(\x_{i}, y_{i}\right)\right\}_{i=1}^{NK}$, where $y_i \in Y_b$, and $Y_b$ is the label space of task $b$. 
$Y_b  \cap Y_{b^\prime} = \varnothing$ for $b\neq b^\prime$. 
We can only access $\D^b$ when training task $b$.
The limited instances in $\D^b$ can be organized as $N$-way $K$-shot data format, \ie, there are $N$ classes in the dataset, and each class has $K$ training images.
Facing a new dataset $\D^b$, a model should learn new classes and meanwhile maintain performance over old classes, \ie, minimize the expected risk $\mathcal{R}(f,b)$ over all the seen classes:
\begin{align} \label{eq:fscil_risk}
	\mathbb{E}_{\left(\x_{j}, {y}_{j}\right) \sim \D_t^{0}\cup \cdots \D_t^{b} }\left[\ell\left(f
	\left(\x_{j}; \D^b, \phi^{b-1}, W^{b-1} 
	\right), {y}_{j}\right)\right] \,.
\end{align}
In Eq.~\ref{eq:fscil_risk}, the learning algorithm $f$ should build the new model based on new dataset $\D^b$ and current old model $\phi^{b-1},W^{b-1}$, and minimize the loss over all seen classes.
Eq.~\ref{eq:fscil_risk} depicts the expected risk with the $b$-th task. Since FSCIL is a dynamic process, the expected risk after \emph{every} incremental session, \ie, $\sum_{b=1}^B \mathcal{R}(f,b)$ should be minimized.

\noindent\textbf{Forgetting in Few-Shot Class-Incremental Learning:} 
An intuitive way for FSCIL is to directly employ CIL algorithms, \eg, knowledge distillation~\cite{hinton2015distilling,li2017learning}. Except for cross-entropy loss $\mathcal{L}_{CE}$, it also builds a mapping between former model and the current model to prevent catastrophic forgetting:
\begin{align} \label{eq:icarl}
	&\mathcal{L}(\x, y)=(1-\lambda) \mathcal{L}_{CE}(\x, y)+\lambda \mathcal{L}_{KD}(\x)\\ \notag
	&	\mathcal{L}_{KD}(\mathbf{x}) =  \sum_{k=1}^{|\mathcal{Y}_{b-1}|}-
	\mathcal{S}_k(\bar{W}^{\top}\bar{\phi}(\x))
	\log \mathcal{S}_k({W}^{\top}\phi(\x)) \,,
\end{align}
where $\mathcal{Y}_{b-1}=Y_0\cup\cdots Y_{b-1}$ denotes the set of old classes, and $\mathcal{S}_k(\cdot)$ denotes the $k$-th class logit after softmax operation. $\bar{W}_{}$ and $\bar{\phi}$ correspond to the frozen classifier and embedding before learning $\D^b$. The knowledge distillation term in Eq.~\ref{eq:icarl} forces the new model to output the same activation result as the old model over seen classes. As a result, the aligned probabilities force the current model to have the same \emph{discrimination} as the old one and prevent the former knowledge from being forgotten. 

\noindent\textbf{Overfitting in Few-Shot Class-Incremental Learning }
Eq.~\ref{eq:icarl} helps to trade off learning new classes and remembering old classes, which works well with \emph{sufficient} instances, \ie, traditional class-incremental scenarios. However, only limited instances are available in FSCIL, making it quickly overfit and destroying the pre-learned embedding. Hence, we need to design and tailor the characteristics of \emph{limited} inputs in FSCIL.
Alternatively, an effective method in the few-shot learning field, \ie, prototypical network~\cite{snell2017prototypical,wang2019simpleshot} shows competitive performance in resist overfitting. It first pre-trains the model on base classes with cross-entropy loss. During incremental sessions, the model fixes the embedding module $\phi(\cdot)$, and extracts the average embedding of each new class as the class prototype:
\begin{align} \label{eq:prototype}
	\p_i=\frac{1}{K}
	{\sum_{j=1}^{|\mathcal{{D}}^b|}\mathbb{I}(y_j=i)\phi(\x_j)}
	\,,
\end{align}
where $\mathbb{I}(\cdot)$ is the indicator function. The class prototype stands for the most representative features of this class, and we can directly substitute the classifier for $i$-th class with the prototype: $\w_i=\p_i$. 
Eq.~\ref{eq:prototype} relies on the generalizability of the embedding module to map new classes with current features.
Since it will not change the pre-trained embedding towards few-shot images, Eq.~\ref{eq:prototype} alleviates the overfitting phenomena.

\noindent\textbf{Learning Generalizable Features}
Eq.~\ref{eq:prototype} adapts to new classes without harming current embedding. However, it only traces the \emph{static} characteristic for new classes, while FSCIL is a \emph{dynamic} process. It is better to consider new classes and build a more general embedding for old and new classes. As a result, we need to design a new learning paradigm that considers the future extending process — if an algorithm can tackle different kinds of simulated FSCIL tasks, it would easily handle the `real' task. 
Besides, simply replacing the classifier with prototypes brings in the semantic gap between the base class classifiers and new class prototypes, which would harm the calibration between old and new classes. Calibration is needed to fill in the semantic gap and encode the inductive bias throughout the training paradigm.

\section{Learning \name for FSCIL}

During the base training session, we can only access base training instances and are not able to optimize the expected risk in Eq.~\ref{eq:fscil_risk} directly. As a result, we need to build a generalizable feature embedding that facilitates \emph{future} few-shot learning with the base instances. Besides, to fill in the semantic gap between new class prototypes and old class classifiers, a proper calibration module should be prepared for calibrating predictions in future incoming tasks. 
Since the target is to optimize the expected risk in Eq.~\ref{eq:fscil_risk} with base classes, an empirical objective consistent with that expected risk shall have better generalizability. 
Taking them into consideration, can we mimic the FSCIL learning scenario in the base session and enable the model to optimize the empirical risk? 
To this end, we sample \emph{sufficient} fake-incremental tasks from the base session, through which we \emph{meta-learn} the ability to handle few-shot class-incremental learning tasks.

We first give our training paradigm and then discuss the design of the meta-calibration module. The deployment guideline for inference is discussed in the last paragraph.

\subsection{Learning Paradigm: Fake-Incremental Tasks}

In few-shot class-incremental sessions, a model should first adapt to new classes with limited instances and then evaluate over all seen classes. As a result, a more generalizable feature will promote the learning of future classes and have better performance in FSCIL. However, we cannot access new classes during the base session, making it impossible to evaluate the quality of the learned embedding. To this end, we are able to sample fake-FSCIL tasks from the base session to empirically search for a `good embedding' facilitating the future learning process. However, the base session contains ample training instances from sufficient classes, which differs from the few-shot input format in the incremental sessions. 
An intuitive idea to unify the base session and incremental session is to  synthesize fake few-shot incremental sessions with \emph{the same data format} as FSCIL and optimize the empirical risk of these fake-tasks. Since the empirical risk shares the same format as Eq.~\ref{eq:fscil_risk}, if the model well tackles these synthesized fake-incremental tasks, such \emph{learning ability} obtained in fake tasks will be generalized to unseen new tasks and perform well.
After that, we can find a suitable embedding that generalizes to incoming new classes.

\noindent\textbf{Fake-task in FSCIL}: To make the fake-incremental tasks share the same data format as `real' FSCIL tasks, we first divide the base label space $Y_0$ into two non-overlapping sets: $Y_0= \hat{Y}_0 \cup \hat{Y}_U$.  $\hat{Y}_0$ stands for the fake-base classes, and $\hat{Y}_U$ stands for the fake-incremental classes. We further divide fake-incremental classes into several $\hat{N}$-way phases to mimic the incremental sessions, \ie, $\hat{Y}_U= \hat{Y}_1 \cup \cdots \hat{Y}_C, |\hat{Y}_c|=\hat{N}$. Without loss of generality, we assume $|\hat{Y}_U|=\hat{N}C$, where $C$ denotes the maximum fake-phase. Afterward, we can randomly sample $\hat{K}$ instances from  the corresponding label space $\hat{Y}_c$ to construct a
$\hat{N}$-way $\hat{K}$-shot fake-training set $\Sup^c$, forming a sequence of support sets $\mathcal{S}^1,\mathcal{S}^2,\cdots,\mathcal{S}^C$. 
Note that we only synthesize few-shot sessions and do not need to sample the many-shot session $\mathcal{S}^0$.
The sampled fake-training set takes the same format as `real' few-shot incremental sessions, and we expect to generalize a good learning strategy $f$ to incoming tasks.  
For example, in fake phase-1, the model needs to classify classes in $\hat{Y}_0 \cup \hat{Y}_1$ based on the training data $\Sup^1$ and the old model trained on $\hat{Y}_0$.
In other words, the divergence  $\ell\left(f
\left(\x_{j}; \Sup^1, \phi^{0}, W^{0} 
\right),{y}_{j}\right)$ should be small for $y_j\in\left(\hat{Y}_0 \cup \hat{Y}_1\right)$.
To empirically evaluate the quality of the learning strategy $f$, we need to prepare `fake-testing sets' to minimize the risk, \ie, Eq.\ref{eq:fscil_risk}.

The difficulty in evaluating $f$ lies in that we do not have `real' FSCIL tasks during meta-training. As a result,
we should design a proper way to evaluate $f$ with the current data, \ie, the base session. To this end, we need to sample an extra `testing' set to empirically analyze the learning strategy's performance after learning fake-incremental tasks.
Recalling that the goal of FSCIL is working well on all \emph{seen} classes, the testing/query set should contain the classes from old and new classes jointly. Correspondingly, we randomly sample the query set $\Q^c$ from all the seen classes $\hat{\mathcal{Y}}_c=\hat{Y}_0 \cup \cdots \hat{Y}_c$, forming a $|\hat{\mathcal{Y}}_c|$-way $K'$-shot dataset, where $K'$ can be large to evaluate the performance holistically. The query set should not contain instances that emerge in the support set, \ie, $\Q^c\cap \Sup^c= \varnothing$, and we can obtain the meta-test datasets: $\mathcal{Q}^1,\mathcal{Q}^2,\cdots,\mathcal{Q}^C$. They are then utilized to evaluate how well the learning algorithm works on the fake-training set $\Sup$. The guidelines for sampling fake incremental tasks are summarized in Algorithm~\ref{alg1:sampling}.

For now, we get the fake-training and fake-testing sets. In each fake-phase, our model  $f$ outputs a $|\hat{\mathcal{Y}}_c|$-way classifier $\hat{W}$ with two steps: (1) for the simulated fake few-shot task $\Sup^c$, it calculates and substitutes the classifier $\w_i$ with prototype $\p_i, \forall i\in \hat{Y}_c$ according to Eq.~\ref{eq:prototype}. (2) for the other old classes, it directly utilizes the classifier at last phase. For the first phase $c=1$, we use the pre-trained weights in
$W_{\hat{Y}_0}$ as the corresponding classifiers:
\begin{align} \label{eq:replace_prototype}
	\hat{W}_i=
	\left\{\begin{array}{l}
		\p_{i},  \quad \quad  \forall i \in \hat{Y}_c \\
		\w_{i}^{c-1},   \quad \forall i \notin \hat{Y}_c 
	\end{array}\right. \,.
\end{align}
 With the help of these sampled fake-incremental tasks, the expected risk over all phases can be substituted into the empirical risk:
	\begin{align} \label{eq:meta-risk} 
	\hat{\mathcal{R}}(f)	 &= \sum_{c=1}^C \hat{\mathcal{R}}(f,c) \,,\\ \notag
	\hat{\mathcal{R}}(f,c)&=	\mathop{\mathbb{E}}_{ \Sup^c,\Q^c\sim  \D^{0} } 
		\mathop{\mathbb{E}}_{\left(\x_{j}, {y}_{j}\right) \in \Q^{c} } 
		\left[\ell\left(f
		\left(\x_{j}; \Sup^c, \phi^{c-1}, W^{c-1} 
		\right), {y}_{j}\right)\right]  
	\end{align}
which can be optimized by sampling sufficient fake tasks from the base session. Note that the loss is measured on the instances from the entire distribution $\Q^{c}$, which contains both base and incremental classes.
 The left part in Figure~\ref{figure:teaser} shows the sampling process, where we can divide the label space and sample instances with humble complexity. Afterward, large amounts of fake tasks are available for optimization, and we can obtain a better embedding space for future tasks.

\noindent\bfname{Summary of fake-incremental tasks:} The core idea of fake-incremental tasks is to synthesize the \emph{long-term} incremental learning process from the base session. Correspondingly, we sample support set $\mathcal{S}$ to mimic the few-shot training session, and sample query  set $\mathcal{Q}$ to act as the testing set to evaluate the learning strategy. Although Eq.~\ref{eq:meta-risk} is optimized over base classes, the encoded incremental learning ability can be generalized to unseen new classes since it has the consistent risk format as Eq.~\ref{eq:fscil_risk}.
Solving Eq.~\ref{eq:meta-risk} can simultaneously differentiate base and new classes, which also trades off among seen classes.
Besides, it reflects the dynamic process of FSCIL with multiple incremental phases.
Take the learning process of a child for an example. Suppose he has different learning habits to recognize new items, and we provide him with various practices to find the most efficient habit. In that case, he shall obtain the learning ability and easily recognize new items in the future.
In the real FSCIL sessions, we substitute $\Sup^c$ into $\D^b$ and get the prediction for testing instances.

	\begin{algorithm}[t]
	\caption{ Sample fake-incremental tasks }
	\label{alg1:sampling}
		{\bf Input}: Base dataset: $\D^0$; fake-phase $C$; fake-way: $\hat{N}$; fake-shot: $\hat{K}$; query shot: $K'$\\
		{\bf Output}: Sampled fake support sets $\Sup^1,\cdots,\Sup^C$ and query sets $\Q^1,\cdots,\Q^C$;
	\begin{algorithmic}[1]{
			\STATE Randomly divide the label space $Y_0= \hat{Y}_0 \cup \hat{Y}_U$ with $|\hat{Y}_U|=\hat{N}C$;
			\STATE Randomly divide the label space $\hat{Y}_U= \hat{Y}_1 \cup \cdots \hat{Y}_C$  with $|\hat{Y}_c|=\hat{N}$;
			\FOR {$i=1,2,\cdots C$}
			\STATE Randomly sample $\hat{K}$ instances per class from the label set $\hat{Y}_i$, forming $\Sup^i$;
			\STATE Randomly sample ${K'}$ instances per class from the label set $\hat{Y}_0\cup\cdots\hat{Y}_i$, forming $\Q^i$;
			\ENDFOR
			\RETURN  $\{\Sup^1,\cdots,\Sup^C, \Q^1,\cdots,\Q^C\}$
		}
	\end{algorithmic}
\end{algorithm}

\subsection{Training Meta-Calibration Module}
During the fake-incremental tasks, the model should adjust itself with the fake-training set $\Sup^c$ and perform well on $\Q^c$. 
As discussed in Eq.~\ref{eq:replace_prototype}, facing an upcoming task, our model  replaces the classifier for the few-shot classes with prototypes $\w_i=\p_i$, and gets the replaced classifier $\hat{W}$.
However, the incremental model is optimized on the many-shot old classes, which is tailored to depict old classes' features. As a result, there exists a \emph{semantic gap} between the old classifiers and extracted new class prototypes, and we would like to derive a calibration between them.
Such a calibration process can also be meta-learned --- if a model calibrates well between prototype and classifiers during the fake-incremental tasks, it will also calibrate well for `real' FSCIL sessions. 
Correspondingly, we design a meta-calibration module, with which we can extract the inductive bias during meta-training and generalize to upcoming incremental sessions.

\noindent\bfname{Characteristics of meta-calibration:} A good calibration module should reflect the \emph{context relationship} between old and new classes.
If the query instance is a `tiger,' then the classifiers and prototypes should be adapted to highlight the discriminative features, \eg, beards and strides. 
If the query instance is a `bird,' then features of wings and beaks should be emphasized. As a result, the calibration process can be seen as conducting `co-adaptation' --- we need to transform the query embeddings and classifiers to highlight the \emph{instance-specific} discriminative features.
Correspondingly, we propose a \emph{set-to-set} function to contextualize query instances and adjusted classifiers. Instance functions cannot reflect the co-adaptation property, which fails to act as the meta-calibration module.  
We denote the adaptation function as $\mathcal{T}(\cdot)$. $\mathcal{T}$ receives classifier and query instance as bags, \ie, $[\hat{W}, \phi(\x)]$, and outputs the set of refined embeddings while being permutation-invariant: $\mathcal{T}\left([\hat{W}, \phi(\x)]\right)=[\tilde{W},\tilde{\phi}(\x)]$.
The predicted logit is calculated as the inner-product of transformed weights and embeddings: $\hat{\mathbf{y}} ={\tilde{W}}^\top \tilde{\phi}(\x)$.
The right part in Figure~\ref{figure:teaser} visualizes $\mathcal{T}$, which inputs the classifiers, prototypes, and the query embedding, and outputs the logit.  
We then introduce the implementation of $\mathcal{T}$.

\noindent\textbf{Implementing $\mathcal{T}$ with Transformer:} 
In our implementation, we utilize Transformer~\cite{vaswani2017attention} architecture to implement the set-to-set function. Specifically, we use the self-attention mechanism~\cite{lin2017structured,vaswani2017attention} to calibrate the query instance with consideration of the old class classifiers and new class prototypes. Transformer architecture is permutation invariant and good at outputting adapted embeddings even with long dependencies, which naturally satisfies the required characteristics of adaptation function $\mathcal{T}(\cdot)$.

Transformer is a store of triplets in the form of (query $\mathcal{Q}$, key $\mathcal{K}$ and value $\mathcal{V}$). Those points are first linearly projected into some space: $K=W_{K}^{\top}\left[\mathbf{k}_{k} ; \forall \mathbf{k}_{k} \in \mathcal{K}\right] \in \mathbb{R}^{d \times|\mathcal{K}|}$.
The projection is the same for $\mathcal{Q}$ and $\mathcal{V}$ with $W_Q$ and $W_V$, respectively.
 It computes what is the right value for a query point — the query $\x_{q} \in \mathcal{Q}$ is first matched against a list of keys $K$ where each key has a value $V$. The final value is then returned as the sum of all the values weighted by the proximity of the key to the query point:
\begin{align}\label{eq:transformer}
\tilde{\phi}(\x)=\phi(\x)+\sum_{k} \alpha_{q k} V_{:, k} \,,
\end{align}
where $
	\alpha_{q k} \propto \exp \left(\frac{\phi(\x)^{\top} W_{Q} \cdot K}{\sqrt{d}}\right) $
and $V_{:, k}$ is the $k$-th column of $V$. The adaptation process is conducted for $\hat{W}$ in the same way.
 In our implementation, to simultaneously emphasize the discriminative features between the query instance and classifiers, we have:
 \begin{align} \label{eq:qkv}
  \mathcal{Q}=\mathcal{K}=\mathcal{V}=[\hat{W}, \phi(\x)] \,.
\end{align}

\begin{table*}[t]
	\centering
	\caption{Ablation study on CIFAR100. The last line stands for \mame. Every part in \name improves the performance of FSCIL.}
	\resizebox{\textwidth}{!}{
		
		\begin{tabular}{cccclllllllllll}
			\toprule
			\multicolumn{1}{c}{\multirow{2}{*}{Prototype}} &
			\multirow{2}{*}{Calibration} &
			\multirow{2}{*}{Meta-1} &
			\multirow{2}{*}{Meta-C} &
			\multicolumn{9}{c}{Accuracy in each session (\%) $\uparrow$} &
			\multirow{2}{*}{PD $\downarrow$} &
			Our relative \\ \cmidrule{5-13}
			\multicolumn{1}{c}{}      &   &   &   & 0     & 1     & 2     & 3    & 4    & 5    & 6    & 7    & 8    &       & improvement               \\ \midrule
			
			&      &      &      & 73.61   & 39.61      & 15.37      & 9.80   & 6.67     & 3.80  & 3.70    & 3.14      & 2.65     & 70.96   & \bf +48.38     \\
			
			\checkmark            &      &      &       & 73.56   & 67.06      & 63.58      & 59.55  & 56.07    & 53.44  & 51.40   & 49.13     & 46.98    & 26.58  & \bf +4.00     \\
			
			\checkmark            &    \checkmark  &      &        & 73.63   & 69.73      & 65.47      & 61.37   &  57.96    &55.09  & 52.90    & 50.53     & 48.30     & 25.33    & \bf +2.75      \\
			
			\checkmark            &    \checkmark  & \checkmark     &           & \bf 74.01  & 70.07      & 66.11      & 61.98   & 59.01    & 56.11  & 54.40    & 52.23      & 50.01    & 24.00  & \bf +1.42      \\

			\midrule
			\checkmark            &    \checkmark  &      & \checkmark  &  73.81   & \bf 72.09    & \bf67.87    & \bf63.89  & \bf60.70   &\bf57.78 & \bf55.68   &\bf 53.56    & \bf51.23   & \bf22.58   &   \\
			
			\bottomrule
		\end{tabular}
	} \label{tab:ablation}
\end{table*}

\begin{figure*}[t]
	\begin{center}
		\subfigure[CIFAR100]
		{\includegraphics[width=.675\columnwidth]{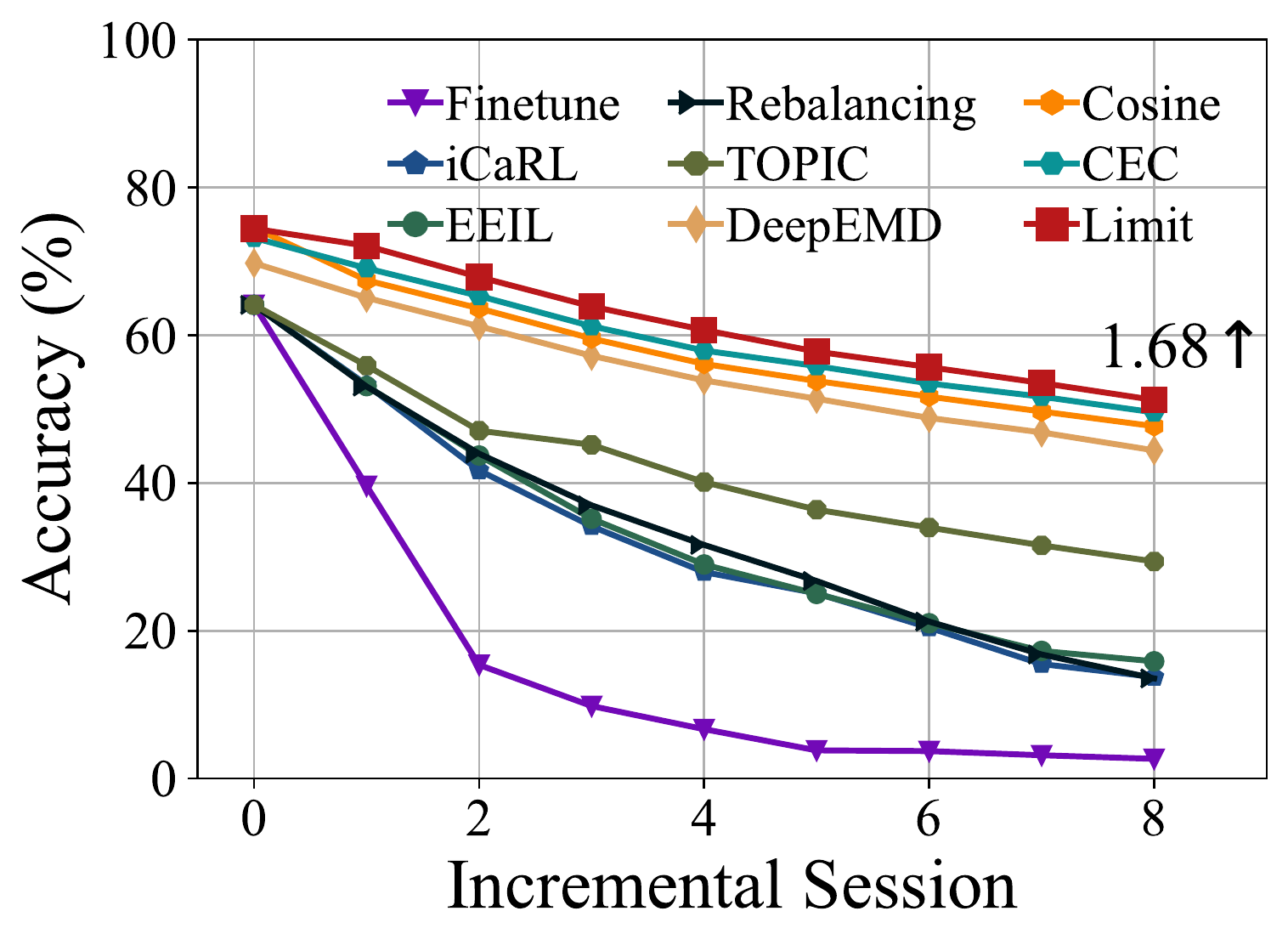}}
		\subfigure[CUB200]
		{\includegraphics[width=.675\columnwidth]{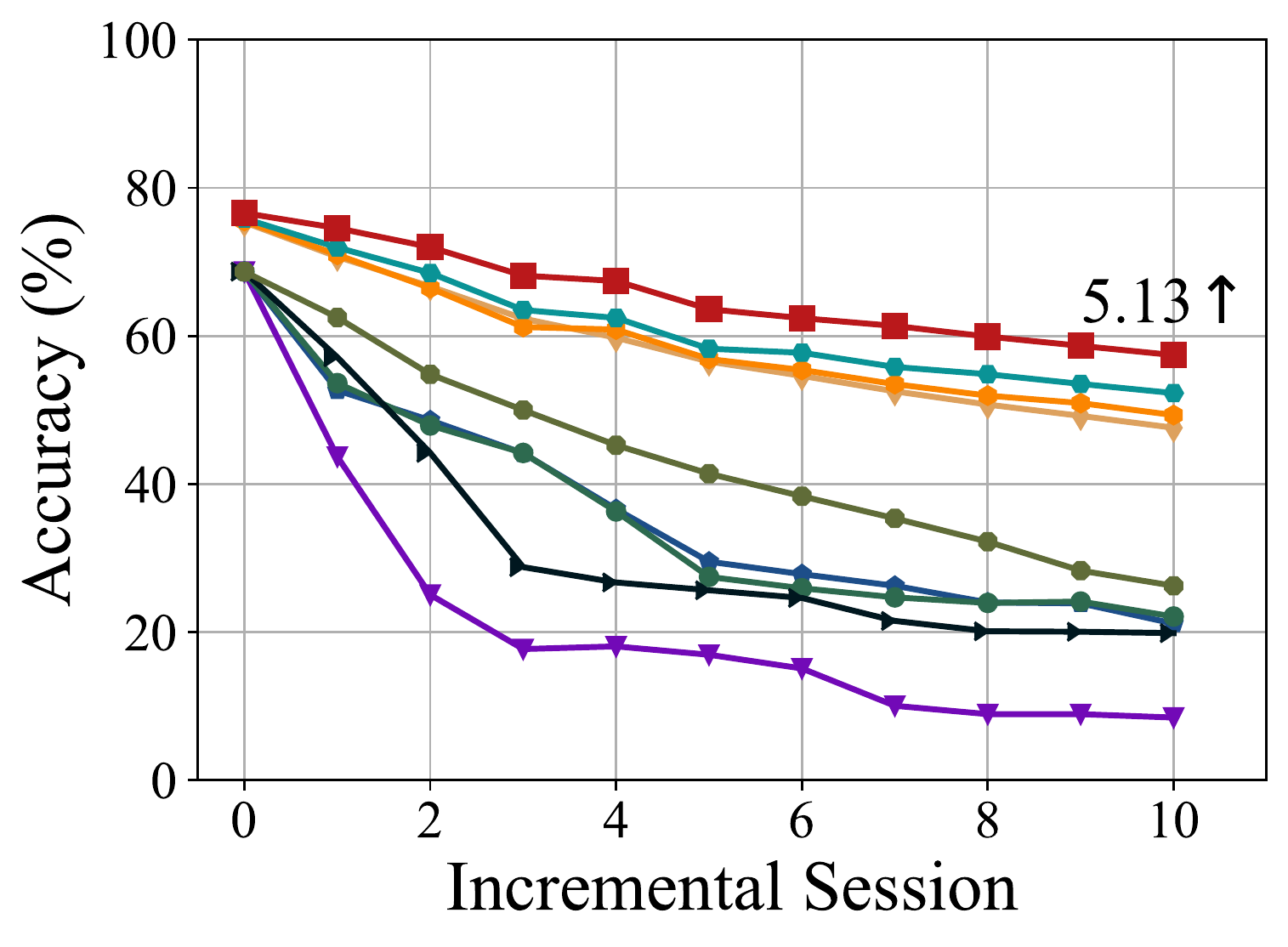}}
		\subfigure[\textit{mini}ImageNet]
		{\includegraphics[width=.675\columnwidth]{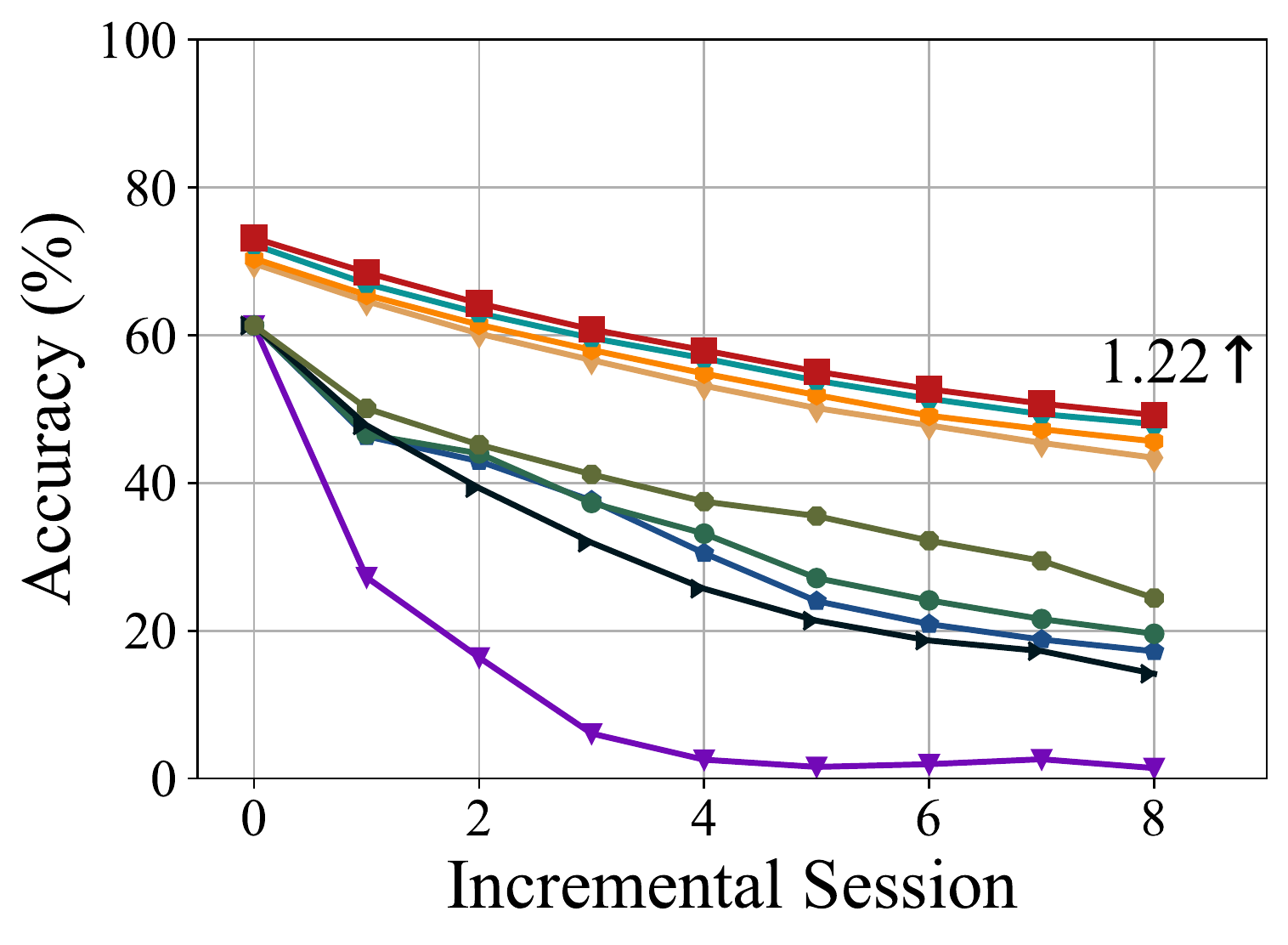}}\\
	\end{center}
	\caption{Top-1 accuracy along few-shot incremental tasks. Legends are shown in (a). We report the performance gap after the last incremental task of \name and the runner-up method at the end of the line. We report the results of CEC with auto augmentation for CIFAR100 and \textit{mini}ImageNet, which works better than basic augmentations. Please refer to
	Table~\ref{table:cub}, \ref{tab:cifar}, \ref{tab:mini} for detailed values.
	} \label{figure:benchmark}
	
\end{figure*}

\noindent\bfname{Effect of meta-calibration:}
With the help of the meta-calibration module, we are able to overcome the obstacle of the semantic gap between old class classifiers and new class prototypes. We can encode the inductive bias of such an adaptation process during meta-training and generalize it to real incremental tasks. Besides, Eq.~\ref{eq:transformer} also provides a way to obtain the instance-specific embedding, which helps to obtain a more discriminative prediction.
 By incorporating the calibration process into meta-training, we can transform the risk in Eq.~\ref{eq:meta-risk} into:
\begin{align} \label{eq:feat}
		\min _{\left\{{\phi},W, \mathcal{T} \right\}}
		\sum_{c=1}^C
		\sum_{\mathcal{S}^c,\mathcal{Q}^c \sim \D^0} \sum_{\left({\x}_{j}, y_{j}\right) \in \mathcal{Q}^c} \ell\left({\tilde{W}}^\top \tilde{\phi}(\x), {y}_{j}\right) \,.
\end{align}
Optimizing Eq.~\ref{eq:feat} helps to extract the inductive bias into $\mathcal{T}$ during  fake-tasks. If the calibration works well for fake tasks, it will generalize to calibrating the `real' FSCIL process.

\noindent\textbf{Summary of meta-calibration:} The ultimate goal of the calibration module is to co-adapt the embedding set and obtain `instance-specific' embedding with context information. Thanks to the fake-incremental learning protocol, the calibration module can be incorporated and optimized during the meta-training process, \ie, in Eq.~\ref{eq:feat}. It should be noted that the transformer structure is a typical format of set-to-set function, and \name is a generalized training protocol for FSCIL. Other formats of set functions that are able to propagate context information among sets can also be applied to \mame, \eg, deep sets~\cite{zaheer2017deep} and bi-directional LSTM~\cite{hochreiter1997long,vinyals2016matching}, which we will
explore in future work.

\begin{algorithm}[t]
	\caption{ Meta-train \name for FSCIL }
	\label{alg1}
	{\bf Input}: Base dataset: $\D^0$; pre-trained model $W,\phi$; fake-phase $C$;\\
	{\bf Output}:
	Optimized model $\hat{W},\hat{\phi}$;  meta-calibration~module~$\mathcal{T}$;
	\begin{algorithmic}[1]{
			\STATE Randomly initialize $\mathcal{T}$;
			\REPEAT
			\STATE Sample $\{\Sup^1,\cdots,\Sup^C;\Q^1,\cdots,\Q^C\}$ with Algorithm~\ref{alg1:sampling}; 
			\FOR {$i=1,2,\cdots C$}
			\STATE Calculate the prototype in $\Sup^i$ by Eq.~\ref{eq:prototype};
			\STATE Replace the classifier with prototypes, get $\hat{W}$;
			\STATE Meta-calibrate and get transformed $\tilde{W}, \tilde{\phi}(\x)$ in Eq.~\ref{eq:transformer};
			\STATE Predict the logits for query instances $\Q^i$: $\hat{\mathbf{y}} ={\tilde{W}}^\top \tilde{\phi}(\x)$;
			\ENDFOR
			\STATE Solve Eq.~\ref{eq:feat}. Obtain derivative and update the model;
			\UNTIL reaches predefined epochs
		}
	\end{algorithmic}
	
\end{algorithm}
\subsection{Pseudo Code and Discussions } \label{sec:discussions}
In the former part, we discuss the training paradigm and meta-calibration module. We first pre-train the model with the base dataset $\D^0$ to get a $|Y_0|$-way classifier and then start meta-training.
 We give the meta-training process of  \name in Algorithm~\ref{alg1}. We first randomly initialize the meta-calibration module (Line 1). In each training iteration, we first sample a sequence of fake-incremental tasks, \ie, $\{\Sup^1,\cdots,\Sup^C;\Q^1,\cdots,\Q^C\}$. During the fake-incremental learning phase, we initialize the classifier for new classes with prototypes and then conduct the calibration process to get the calibrated logits and optimize Eq.~\ref{eq:feat}. We finish meta-training when it reaches the predefined epochs (Line 11). Afterward, the model is well-prepared for the real FSCIL sessions.

\noindent\textbf{Inference guidelines:}
Facing the incoming FSCIL sessions, \eg, $\D^b$, we need to calculate the prototype in $\D^b$, and augment the current classifier with new class prototypes $[W,\p_{j}, \forall j \in {Y}_b]$. Then the model updating process is finished. During inference time, the same calibration process in Line 7-8 will be deployed to get the prediction on testing sets. The updating process of these `real' incremental tasks is the same as the meta-training tasks. Thus, the generalizable feature and calibration  information extracted from the meta-training phase can be generalized to the real FSCIL tasks. The aforementioned model updating process does not involve the back-propagation, and the model is updated by solely augmenting the classifiers with prototypes in incremental sessions.

Note that there is no requirement of the consistency between meta-training and real incremental learning, \ie, we do not require $\hat{N}=N$ or $\hat{K}=K$. Firstly, the prototype calculation process (Eq.~\ref{eq:prototype}) can be applied to any-shot instances, and the number of instances per class will not influence the training/testing protocol. On the other hand, the classifier augmentation process $[W,\p_{j}, \forall j \in {Y}_b]$ will include new classes from the set ${Y}_b$, no matter how many classes in it. Besides, the set-to-set function is inherently expandable, and the self-attention calculation can apply to an increasing number of inputs (classifiers).
 Hence, \name can be applied to `any-way any-shot' FSCIL problem without adjustment.

\noindent\bfname{Discussion about related work:} CEC~\cite{zhang2021few} tackles the FSCIL task by designing continually evolved classifiers, which also addresses training pseudo incremental learning. \cite{zhu2021self} proposes a similar training paradigm called random episode selection. Different from CEC, our sampling framework is directly derived from the expected risk in Eq.~\ref{eq:fscil_risk}. As a result, \name optimizes the multi-phase risk by solving Eq.~\ref{eq:meta-risk}, while CEC only minimizes the \emph{approximation} with `one-stage' tasks. 
 Sampling one-stage fake-incremental learning task is a particular case and degradation version for multiple phases. However, the real incremental models will face multiple incremental sessions instead of a single session.
To make training and testing under a \emph{consistent} scheme, we argue that sampling multi-phase fake-tasks is better than one-phase and more suitable for FSCIL training.
We also verify such claims in the experimental verification in Section~\ref{sec:hyperparam}.
Besides, different from the image rotation process in CEC, our task sampling process does not require extra matrix calculation, which is more time-efficient. Additionally, the rotated instances are irrelevant to the FSCIL context, while our sampling strategy is relevant to it and will not harm the semantic information of instances.
 It should be noted that CEC utilizes graph attention network to adjust classifiers during model updating, which is instance-agnostic. However, in \mame, the adjust process happens during the inference, which helps contextualize the \emph{instance-specific} embedding. We  empirically evaluate these different fake task sampling strategies in Section~\ref{sec:c2cec}.

\section{Experiment}
In this section, we compare \name on benchmark and large-scale few-shot class-incremental learning datasets with state-of-the-art methods. Ablations show the effectiveness of sampling fake-incremental tasks and training meta-calibration module, and visualizations indicate \mame's ability to update with limited instances. We also analyze the influence of hyper-parameters in fake-incremental learning on the ability to learn new classes.

\subsection{Implementation Details}
\noindent {\bf Dataset}: Following the benchmark setting ~\cite{tao2020few}, we evaluate the performance on CIFAR100~\cite{krizhevsky2009learning}, CUB200-2011~\cite{WahCUB2002011} and {\it mini}ImageNet~\cite{russakovsky2015imagenet}. Additionally, we also compare the learning performance on large-scale dataset, \ie, ImageNet ILSVRC2012~\cite{deng2009imagenet}.  They are listed as:

\begin{table*}[t] 
	\caption{ Comparison with the state-of-the-art on CUB200 dataset. We report the results of compared methods from~\cite{tao2020few} and~\cite{zhang2021few}. 
		\name outperforms the runner-up method by 5.13 for the last accuracy and 5.09 for the performance dropping rate. 	\mame$^\dagger$ denotes our method with the same data augmentation (random crop and random horizontal flip) in CEC.
	}
	\centering	
	\resizebox{\textwidth}{!}{
		
		\begin{tabular}{llllllllllllll}
			\toprule
			\multicolumn{1}{c}{\multirow{2}{*}{Method}} & \multicolumn{11}{c}{Accuracy in each session (\%) $\uparrow$} & \multirow{2}{*}{PD $\downarrow$} & {Our relative
			} \\ \cmidrule{2-12}
			\multicolumn{1}{c}{} & 0   & 1      & 2      & 3    & 4     & 5  & 6     & 7      & 8     &   9  & 10   & & improvement    \\ \midrule
			Finetune                & 68.68   & 43.70      & 25.05    & 17.72   & 18.08     & 16.95  & 15.10  & 10.06     & 8.93   &8.93  & 8.47   & 60.21&\bf +41.73     \\
			iCaRL~\cite{rebuffi2017icarl}        & 68.68   & 52.65     & 48.61    & 44.16  & 36.62   & 29.52  & 27.83  & 26.26    & 24.01   &23.89  & 21.16  & 47.52&\bf +29.04   \\
			EEIL~\cite{castro2018end}         & 68.68   & 53.63      &47.91    & 44.20  & 36.30     & 27.46  & 25.93  & 24.70     & 23.95   &24.13 & 22.11  & 46.57&\bf +28.09    \\
			Rebalancing~\cite{hou2019learning}         & 68.68   & 57.12     & 44.21    & 28.78  & 26.71    & 25.66  & 24.62  & 21.52    & 20.12   &20.06  & 19.87  & 48.81&\bf +30.33     \\
			TOPIC~\cite{tao2020few}            & 68.68   & 62.49      & 54.81    & 49.99   & 45.25     & 41.40 & 38.35  & 35.36    & 32.22  &28.31 & 26.26   & 42.40&\bf +23.92      \\
			Decoupled-Cosine~\cite{vinyals2016matching}    &75.52  & 70.95     & 66.46   & 61.20   & 60.86    & 56.88  & 55.40  & 53.49     & 51.94   &50.93 & 49.31   & 26.21&\bf +7.73     \\
			Decoupled-DeepEMD~\cite{zhang2020deepemd}     & 75.35   & 70.69     & 66.68   & 62.34  & 59.76     & 56.54  & 54.61  & 52.52    &50.73   &49.20 & 47.60   & 27.75&\bf +9.27    \\
			CEC~\cite{zhang2021few}                & 75.85   & 71.94    & 68.50   & 63.50   & 62.43    & 58.27 & 57.73 & 55.81    &54.83  &53.52  & 52.28   & 23.57&\bf +5.09   \\
			CEC~\cite{zhang2021few} + AutoAug                & 75.74   & 71.77    & 68.46   & 63.21   & 61.95    & 57.44 & 56.97 & 55.24    &54.23  &52.95  & 51.38   & 24.36&\bf +5.88   \\
			\midrule
			\mame$^\dagger$             &  76.32   &  74.18     & 72.68    & 69.19  & 68.79   &65.64  & 63.57   & 62.69     & 61.47   & 60.44   & 58.45& 17.87 \\
			\name             & \bf 75.89   & \bf 73.55     & \bf71.99    & \bf68.14  & \bf67.42   &\bf63.61  & \bf62.40   &\bf 61.35     & \bf59.91   & \bf58.66   & \bf57.41&\bf 18.48 \\

			\bottomrule
		\end{tabular}
	}\label{table:cub}
\end{table*}

\begin{table*}[t]
	\centering{
		\caption{Comparison with the state-of-the-art on CIFAR100 dataset. We report the results of compared methods from~\cite{tao2020few} and~\cite{zhang2021few}.  
			\name outperforms the runner-up method by 2.09 for the last accuracy and 1.35 for the performance dropping rate.  \mame$^\dagger$ denotes our method with the same data augmentation (random crop and random horizontal flip) in CEC.
		}\label{tab:cifar}
		\resizebox{\textwidth}{!}{
			\begin{tabular}{llllllllllll}
				\toprule
				\multicolumn{1}{c}{\multirow{2}{*}{Method}} & \multicolumn{9}{c}{Accuracy in each session (\%) $\uparrow$} & \multirow{2}{*}{PD $\downarrow$} & {Our relative
				} \\ \cmidrule{2-10}
				\multicolumn{1}{c}{} & 0   & 1      & 2      & 3    & 4     & 5  & 6     & 7      & 8     &     &  improvement      \\ \midrule
				Finetune                & 64.10   & 39.61      & 15.37      & 9.80   & 6.67     & 3.80  & 3.70    & 3.14      & 2.65     & 61.45   & \bf +38.87     \\
				iCaRL~\cite{rebuffi2017icarl}       & 64.10   & 53.28      & 41.69      & 34.13   & 27.93     & 25.06  & 20.41   & 15.48      &13.73    & 50.37  & \bf +27.79  \\
				EEIL~\cite{castro2018end}         & 64.10   & 53.11     & 43.71     & 35.15   & 28.96     & 24.98  & 21.01    & 17.26     & 15.85    & 48.25  & \bf +25.67    \\
				Rebalancing~\cite{hou2019learning}           & 64.10   & 53.05     & 43.96      & 36.97   & 31.61     & 26.73  & 21.23   & 16.78    & 13.54     &50.56  & \bf +27.98   \\
				TOPIC~\cite{tao2020few}                & 64.10   & 55.88     & 47.07      & 45.16   & 40.11   & 36.38 & 33.96   & 31.55      & 29.37     & 34.73   & \bf +12.15      \\
				
				Decoupled-Cosine~\cite{vinyals2016matching}  & 74.55   & 67.43      & 63.63      & 59.55  & 56.11    & 53.80  & 51.68   & 49.67     & 47.68     & 26.87  & \bf +4.29     \\
				Decoupled-DeepEMD~\cite{zhang2020deepemd}    & 69.75   & 65.06     & 61.20     & 57.21  & 53.88    & 51.40  & 48.80  & 46.84     & 44.41     & 25.34   & \bf +2.76    \\
				CEC~\cite{zhang2021few}                    & 73.07   & 68.88     & 65.26    & 61.19  & 58.09   &55.57  & 53.22   & 51.34     & 49.14   & 23.93   & \bf +1.35    \\
				CEC~\cite{zhang2021few}+ AutoAug      & 73.17   & 69.06    & 65.31    & 61.21  & 57.92   &55.82  & 53.47   & 51.67     & 49.55   & 23.62   & \bf +1.04    \\
				\midrule
				\mame$^\dagger$                 & 73.02   & 70.76     & 67.45    & 63.38  & 59.97   &56.90  & 54.84   & 52.18     & 49.92   & 23.10  &    \\
				\name        & \bf 73.81   & \bf 72.09    & \bf67.87   & \bf63.89 & \bf60.70  &\bf57.77 & \bf55.67   &\bf  53.52     & \bf51.23   & \bf22.58   &   \\

				\bottomrule
			\end{tabular}
	}}
\end{table*}

\begin{table*}[t]
	\centering{
		\caption{Comparison with the state-of-the-art on \textit{mini}ImageNet dataset. We report the results of compared methods from~\cite{tao2020few} and~\cite{zhang2021few}. 
			\name outperforms the runner-up method by 1.56 for the last accuracy and 1.24 for the performance dropping rate. 
		\mame$^\dagger$ denotes our method with the same data augmentation (random crop and random horizontal flip) in CEC. }\label{tab:mini}
		\resizebox{\textwidth}{!}{
			
			\begin{tabular}{llllllllllll}
				\toprule
				\multicolumn{1}{c}{\multirow{2}{*}{Method}} & \multicolumn{9}{c}{Accuracy in each session (\%) $\uparrow$} & \multirow{2}{*}{PD $\downarrow$} & {Our relative
				} \\ \cmidrule{2-10}
				\multicolumn{1}{c}{} & 0   & 1      & 2      & 3    & 4     & 5  & 6     & 7      & 8     &     &  improvement      \\ \midrule
				Finetune                & 61.31  & 27.22      & 16.37     & 6.08   & 2.54     & 1.56  & 1.93    & 2.60      & 1.40     & 59.91   & \bf +36.78    \\
				iCaRL~\cite{rebuffi2017icarl}       & 61.31   & 46.32     & 42.94      & 37.63   & 30.49     & 24.00  & 20.89   & 18.80      &17.21  & 44.10  & \bf +20.97 \\
				EEIL~\cite{castro2018end}         & 61.31   & 46.58     & 44.00    & 37.29   & 33.14    & 27.12  & 24.10    & 21.57     & 19.58    & 41.73 & \bf +18.60   \\
				Rebalancing~\cite{hou2019learning}           & 61.31   & 47.80     & 39.31      & 31.91  & 25.68     & 21.35  & 18.67   & 17.24    & 14.17     &47.14  & \bf +24.01    \\
				TOPIC~\cite{tao2020few}                & 61.31 & 50.09    & 45.17      & 41.16   & 37.48   & 35.52 & 32.19  & 29.46    & 24.42     & 36.89   & \bf +13.76      \\
				
				Decoupled-Cosine~\cite{vinyals2016matching}  & 70.37  & 65.45      & 61.41     & 58.00  & 54.81   & 51.89  & 49.10   & 47.27     & 45.63    & 24.74 & \bf +1.61   \\
				Decoupled-DeepEMD~\cite{zhang2020deepemd}    & 69.77   & 64.59     & 60.21    & 56.63  & 53.16   & 50.13  & 47.79  & 45.42     & 43.41    & 26.36   & \bf +3.23   \\
				CEC~\cite{zhang2021few}                    & 72.00   & 66.83     & 62.97   & 59.43  & 56.70   &53.73  & 51.19   & 49.24     & 47.63   & 24.37   & \bf +1.24     \\
				CEC~\cite{zhang2021few} + AutoAug                    & 72.23   & 66.96     & 62.98   & 59.62  & 56.86   &53.85  & 51.40   & 49.32     & 47.97   & 24.26   & \bf +1.13     \\
				\midrule
				\mame$^\dagger$                & 71.92   & 67.93     & 63.58   & 60.22  & 57.33   &54.30  & 51.97   & 50.01     & 48.40   & 23.52   &     \\
				\name              & \bf 72.32   & \bf 68.47     & \bf64.30    & \bf60.78  & \bf57.95   &\bf55.07  & \bf52.70  &\bf 50.72     & \bf 49.19   & \bf23.13   &   \\
				\bottomrule
				
			\end{tabular}
	}}
\end{table*}

\begin{itemize}
	\item{  \bfname{CIFAR100}}: There are 100 classes with 60,000 images. 
	Following the dataset splits of~\cite{tao2020few}, 60 classes are used as base classes, and the rest 40 classes are organized as incoming new classes. We then divide these classes into eight incremental sessions; each contains a 5-way 5-shot incremental task.

	\item \bfname{CUB200}: a fine-grained image classification task with 11,788 images from 200 classes. We follow the dataset configuration in~\cite{tao2020few} and use 100 classes as base classes. The other 100 classes are formulated into ten sessions; each contains a 10-way 5-shot incremental task.
	
	\item{\bfname{{\textit {mini}}ImageNet}}:  is a subset of ImageNet~\cite{deng2009imagenet} with 100 classes. 
	These classes are divided into 60 base classes and 40 new classes. New classes are organized into eight sessions, and each contains a 5-way 5-shot incremental task.
	
	\item{\bfname{ImageNet1000}}:  is a large scale dataset with 1,000 classes, with about
	1.28 million images for training and 50,000 for validation. Following the split of CIFAR100, we divide these classes into 600 base classes and 400 new classes. The new classes are organized into eight sessions, and each contains a 50-way 5-shot incremental task. We also follow~\cite{yu2020semantic} to sample a 100 class subset of ImageNet1000, denoted as \textbf{ImageNet100}. The dataset split of ImageNet100 is the same as CIFAR100.

\end{itemize}

We use the \emph{same} training splits (including instances of base and incremental sessions) for every compared method for a fair comparison. The testing set is the same as the original one to evaluate the generalization ability holistically. We use the policies in AutoAugment~\cite{cubuk2019autoaugment} for data augmentations, \ie, {\it CIFAR10 policy} for CIFAR100, and {\it ImageNet policy} for other datasets. Please refer to Section~\ref{sec:data_aug} for more details.

\begin{figure}[t]
	\begin{center}
		\subfigure[ImageNet100]
		{	\includegraphics[width=.475\columnwidth]{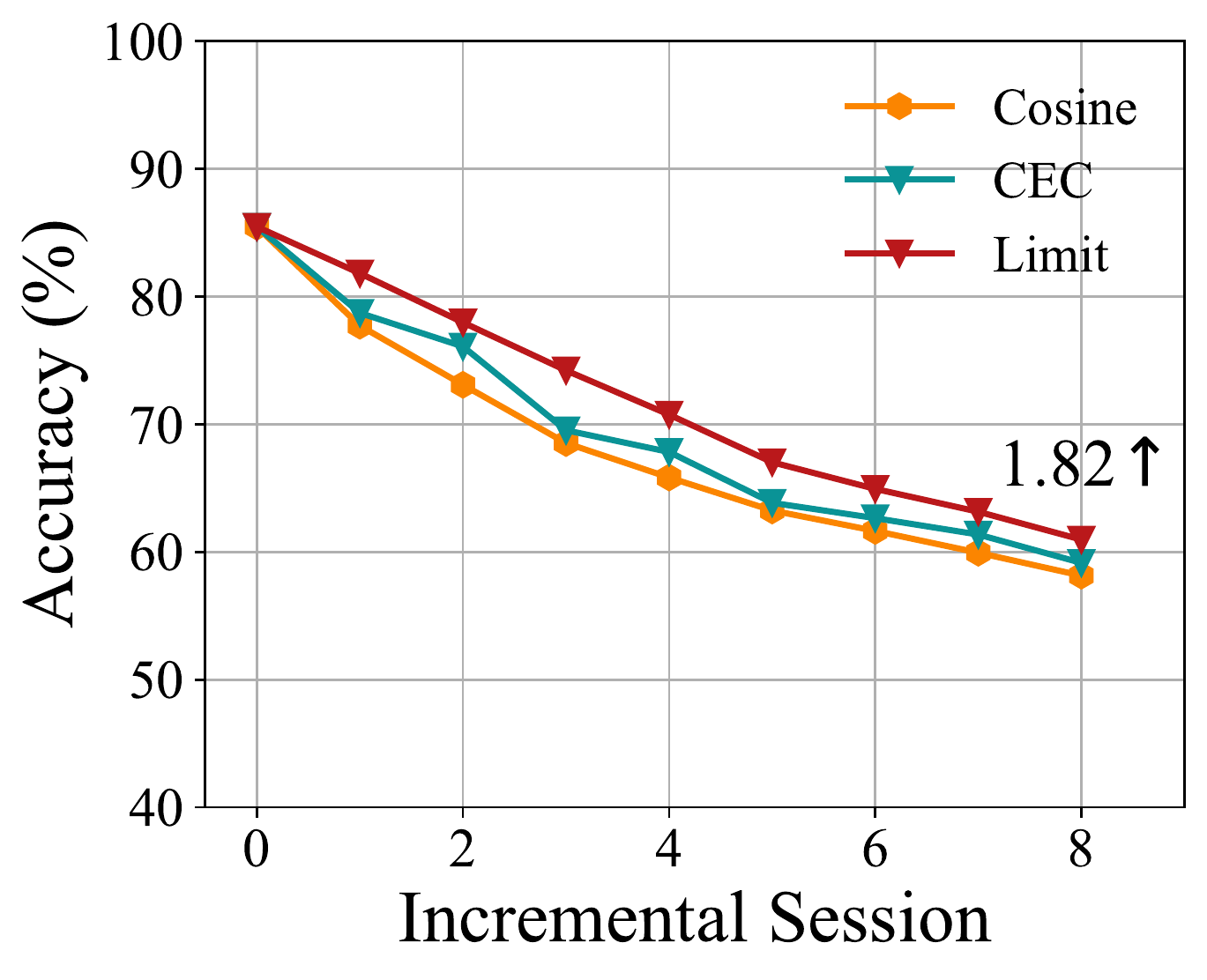}}
		\subfigure[ImageNet1000]
		{		\includegraphics[width=.475\columnwidth]{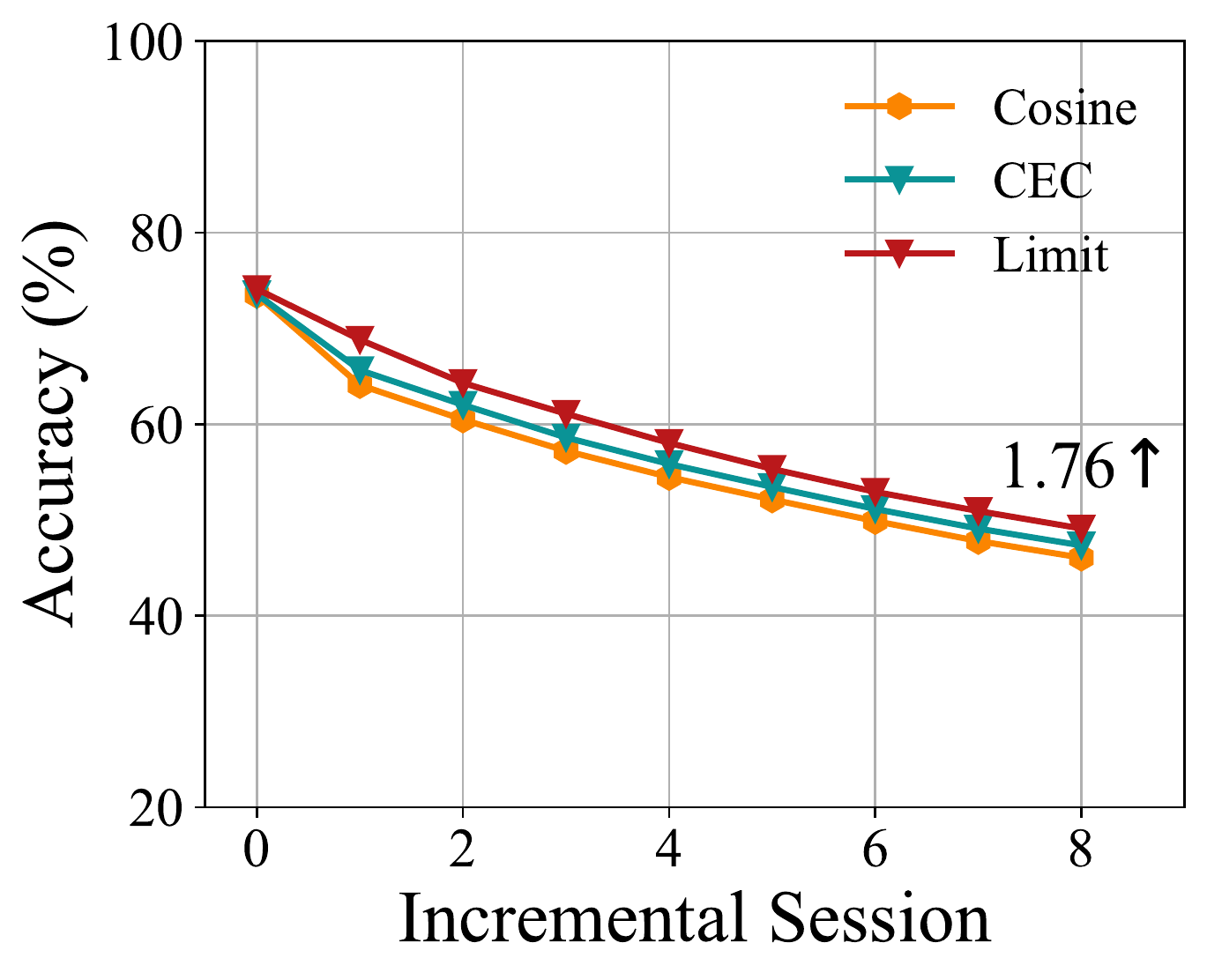}}
		
	\end{center}
	\caption{ Incremental accuracy along incremental tasks on ImageNet. \name consistently outperforms the state-of-the-art method with a substantial margin.
	} \label{figure:imagenet}
\end{figure}

\begin{figure*}[h]
	\begin{center}
		\subfigure[Finetune]{
			\includegraphics[width=.465\columnwidth]{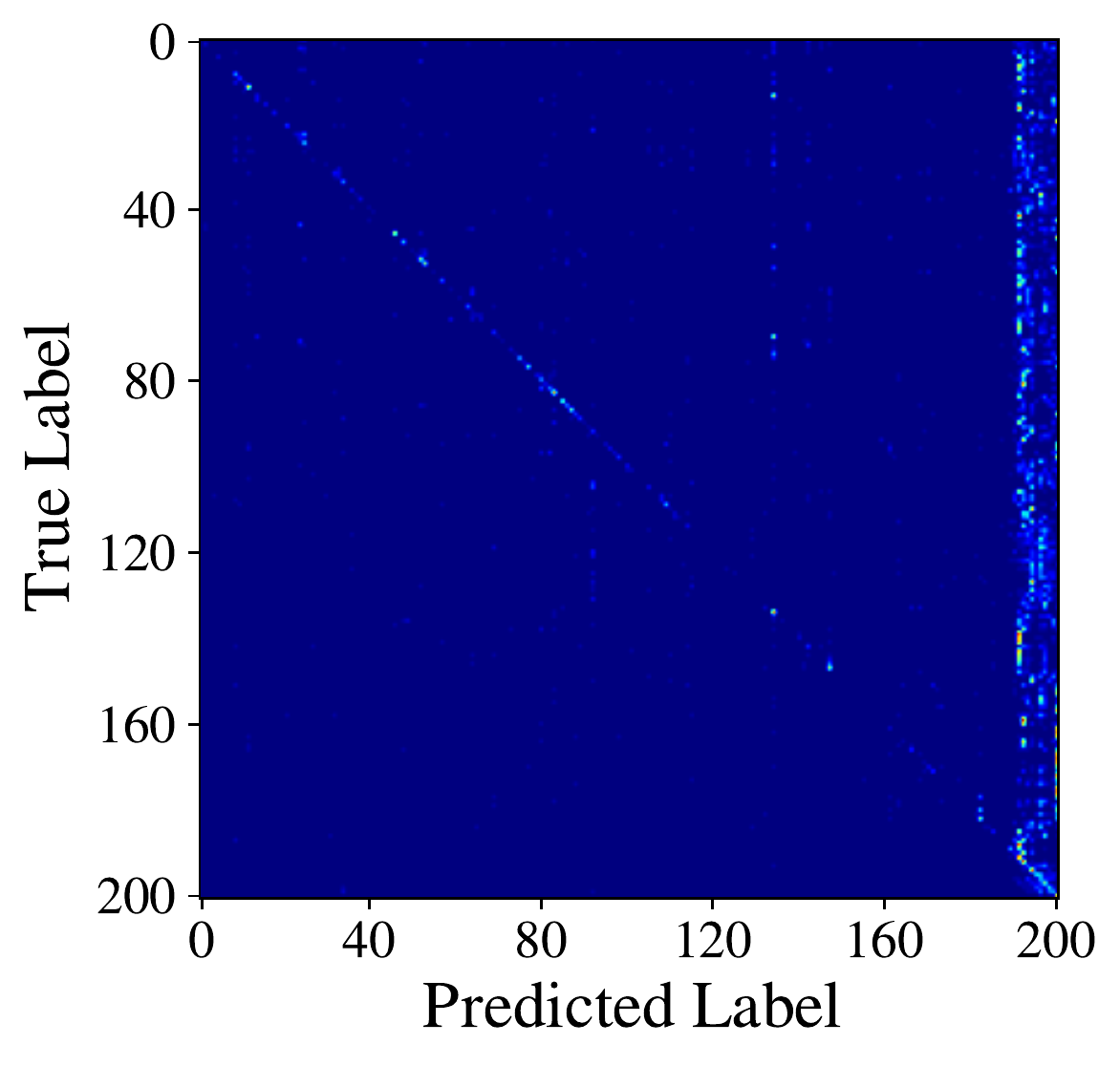}
			\label{fig:ft}}
		\subfigure[iCaRL]{
			\includegraphics[width=.465\columnwidth]{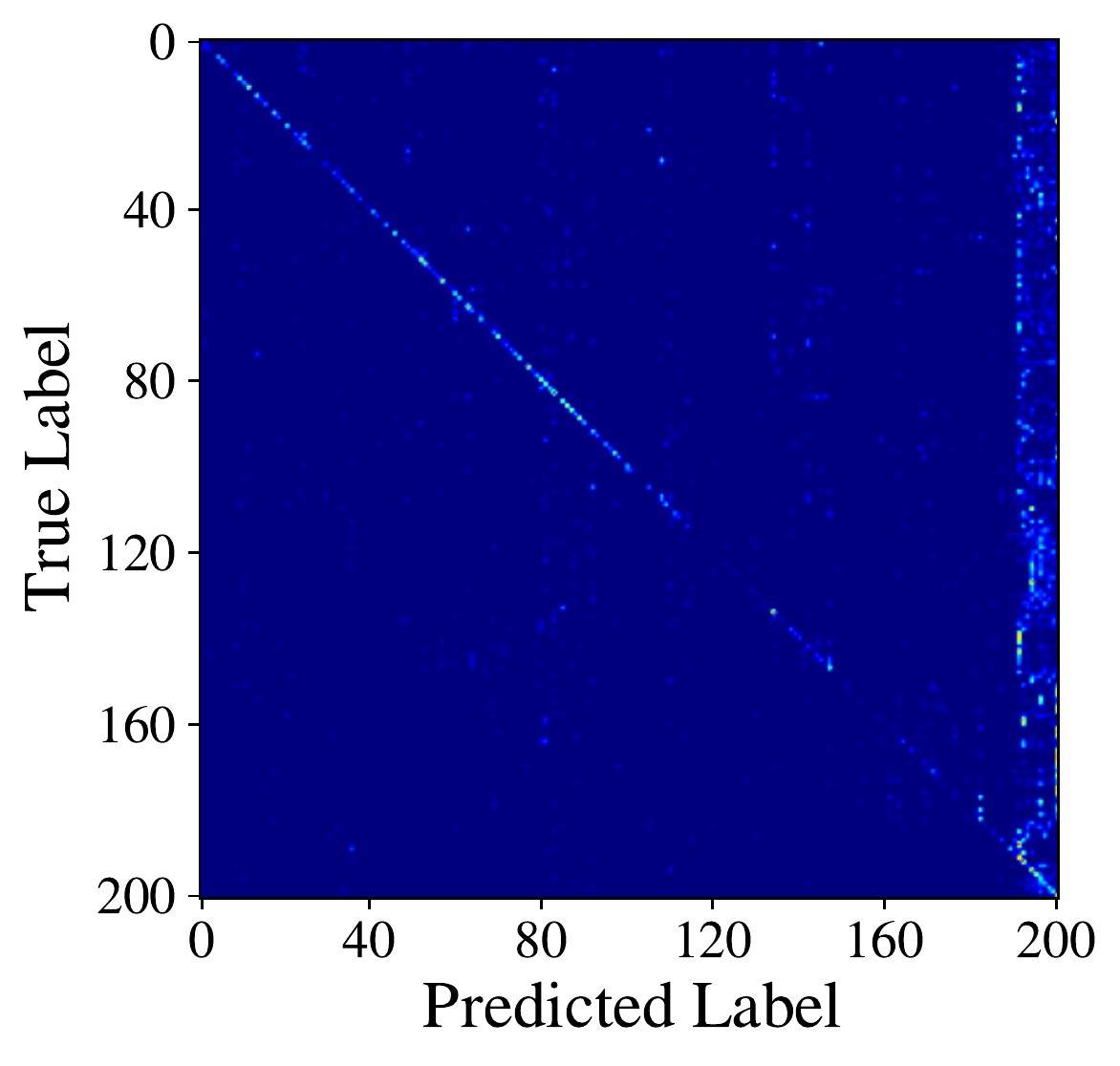}
			\label{fig:icarl}}
		\subfigure[Decoupoled-Cosine]{
			\includegraphics[width=.465\columnwidth]{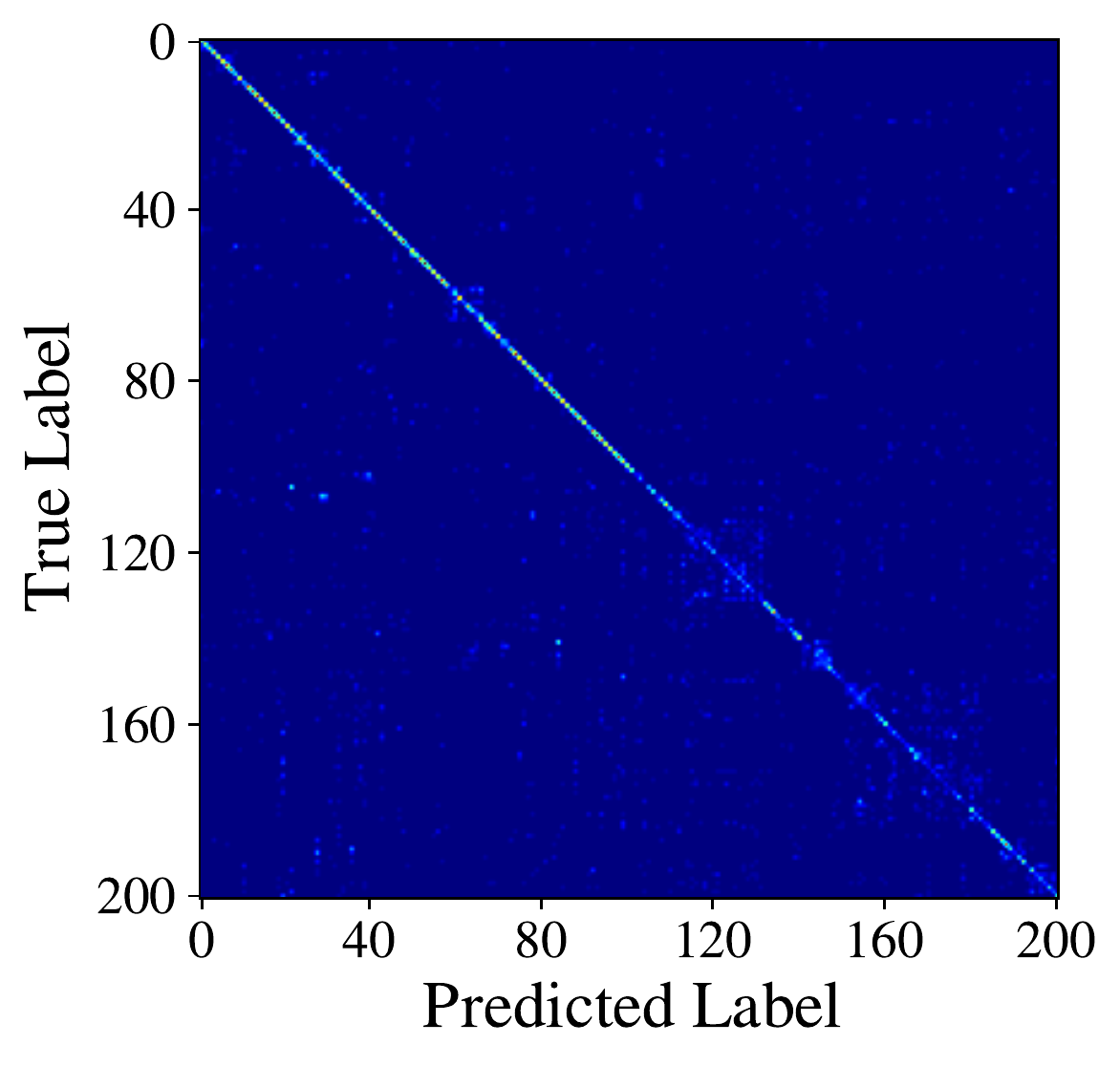}
			\label{fig:cosine}}
		\subfigure[\name]{
			\includegraphics[width=.536\columnwidth]{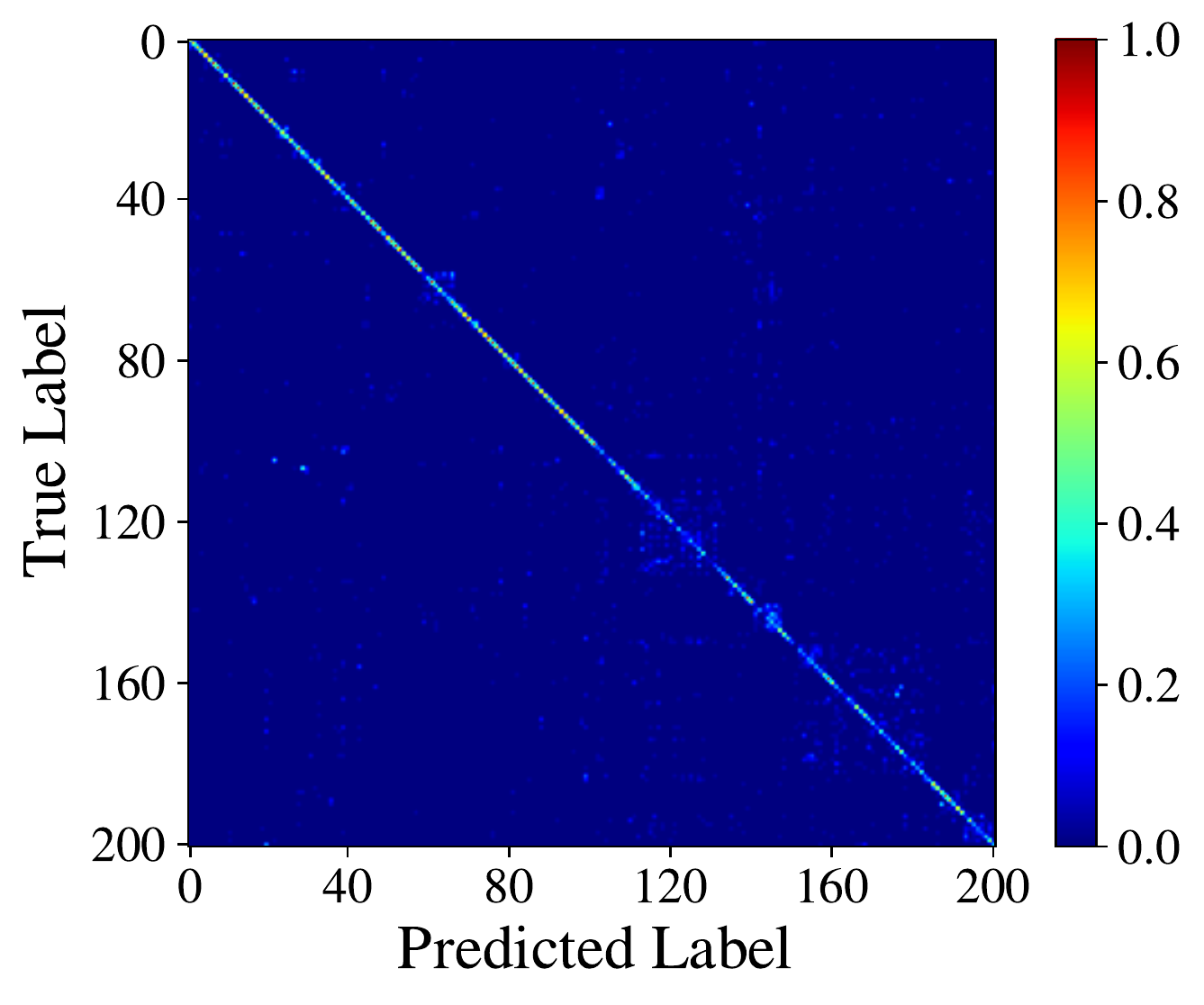}
			\label{fig:ours}}
	\end{center}
	
	\caption{  Confusion matrix on CUB200 after the last incremental session.  	{\name adapts to new classes with a generalizable feature and stably resists catastrophic forgetting. }
	} \label{figure:confmat}
	
\end{figure*}

\noindent {\bf Compared methods:} We first compare to classical class-incremental learning methods, \eg, iCaRL~\cite{rebuffi2017icarl}, EEIL~\cite{castro2018end} and Rebalancing~\cite{hou2019learning}. Besides, we also compare to current state-of-the-art FSCIL algorithms: TOPIC~\cite{tao2020few}, 
 Decoupled-DeepEMD/Cosine~\cite{zhang2020deepemd,vinyals2016matching} and CEC~\cite{zhang2021few}. We also report the baseline  which finetunes the limited instances, denoted as finetune.

\noindent {\bf Training details:} 
All models are deployed with PyTorch~\cite{paszke2019pytorch}.
 We use the \emph{same} network backbone~\cite{tao2020few} for \emph{all} compared methods. For CIFAR100, we use ResNet20~\cite{he2015residual}, while for CUB200, ImageNet and {\it mini}ImageNet we use ResNet18. We follow the most standard implementations for transformer~\cite{vaswani2017attention}. We use a shallow transformer with only one layer, and the number of multi-head attention is set to 1 in our implementation.  The hidden dimension is set to 64 for CIFAR100 and 512 for CUB, ImageNet and \textit{mini}ImageNet. The dropout rate in transformer is set as 0.5.
 Before the fake-FSCIL learning process, we pre-train the model using cross-entropy loss. We use SGD with an initial learning rate of 0.1 and momentum of 0.9. The pre-training epoch is set to 300 for all datasets with a batch size of 128. After that, the model is utilized as the initialization of fake-incremental learning. The learning rate is set to $0.0002$ and suffers a decay of 0.5 every 1000 iterations.
 During the fake-FSCIL learning process, we sample 2-phase fake-tasks to optimize the model. 
 The fake-shot and fake-way are discussed in Section~\ref{sec:hyperparam}. The backbone and the calibration module are fixed after the meta-training process.
  The source code of \name will be made publicly available upon acceptance.

\noindent {\bf Evaluation protocol:}  Following~\cite{tao2020few,zhang2021few}, we denote the Top-1 accuracy after the i-th session as $\mathcal{A}_i$. Algorithms with higher $\mathcal{A}_i$ have the better prediction accuracy.
To quantitatively evaluate the forgetting phenomena of each method, we also use performance dropping rate (PD), \ie, $\text{PD}=\mathcal{A}_0-\mathcal{A}_B$, where $\mathcal{A}_0$ stands for the accuracy after the base session, and $\mathcal{A}_B$ is the accuracy after the last incremental session. The method with a lower performance dropping rate suffers less forgetting phenomena.

\experimentsection{Ablation Study}
We first analyze the importance of each component in \name on the CIFAR100 dataset. The results are shown in Table~\ref{tab:ablation}. We separately construct models with different combinations of the core elements in \mame, \eg, replace classifier with $\p_i$ (denoted as `Prototype'), utilize meta-calibration module $\mathcal{T}$ for calibration (denoted as `Calibration'), train with one-phase fake-tasks (denoted as `Meta-1', \ie, the degradation version of \mame), train with multi-phase fake-tasks (denoted as `Meta-C'. We report the results for $C=2$).

From Table~\ref{tab:ablation}, we can infer that directly optimizing the model with few-shot images, \ie, without any elements in \name will suffer severe catastrophic forgetting (Line 1). Besides, utilizing the prototypes to initialize new class classifiers can relieve the overfitting and forgetting phenomena to some extent (Line 2). However, training the meta-calibration module can further calibrate the relationship between old and new classes, which improves the performance (Line 3). It should be noted that the transformer structure is not used as the embedding in our model, and the main reason for the performance improvement is the learned calibration information during meta-learning.
 Additionally, when using our fake-incremental training scheme, the prediction performance will be further improved (Line 4 and 5). When comparing Line 5 to Line 4, we can infer that multi-phase meta-training helps the model prepare for the multi-phase incremental training process and helps obtain a more generalizable feature space.  
 These results imply that our proposed training paradigm obtains \emph{substantial improvement over the baseline method, \ie, ProtoNet} (Line 2). Besides, learning the meta-calibration module and the meta-learning protocol is helpful for the FSCIL task.

\begin{figure*}[t]
	\begin{center}
		
		\subfigure[Fake-way/shot ]
		{	\includegraphics[width=.64\columnwidth]{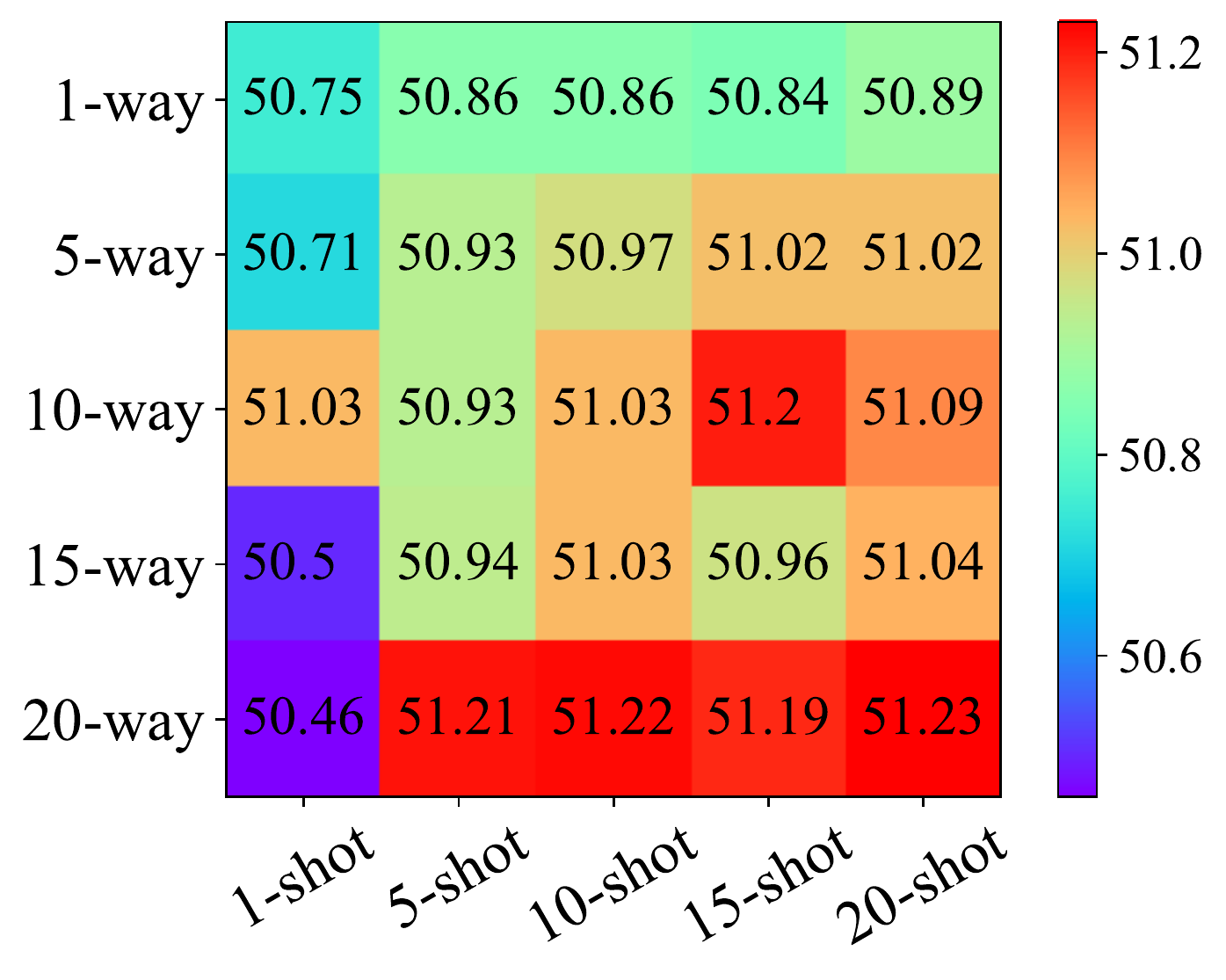}
			\label{fig:cifar-wayshot}}
		\hfill
		\subfigure[Fake-incremental phase]
		{	\includegraphics[width=.64\columnwidth]{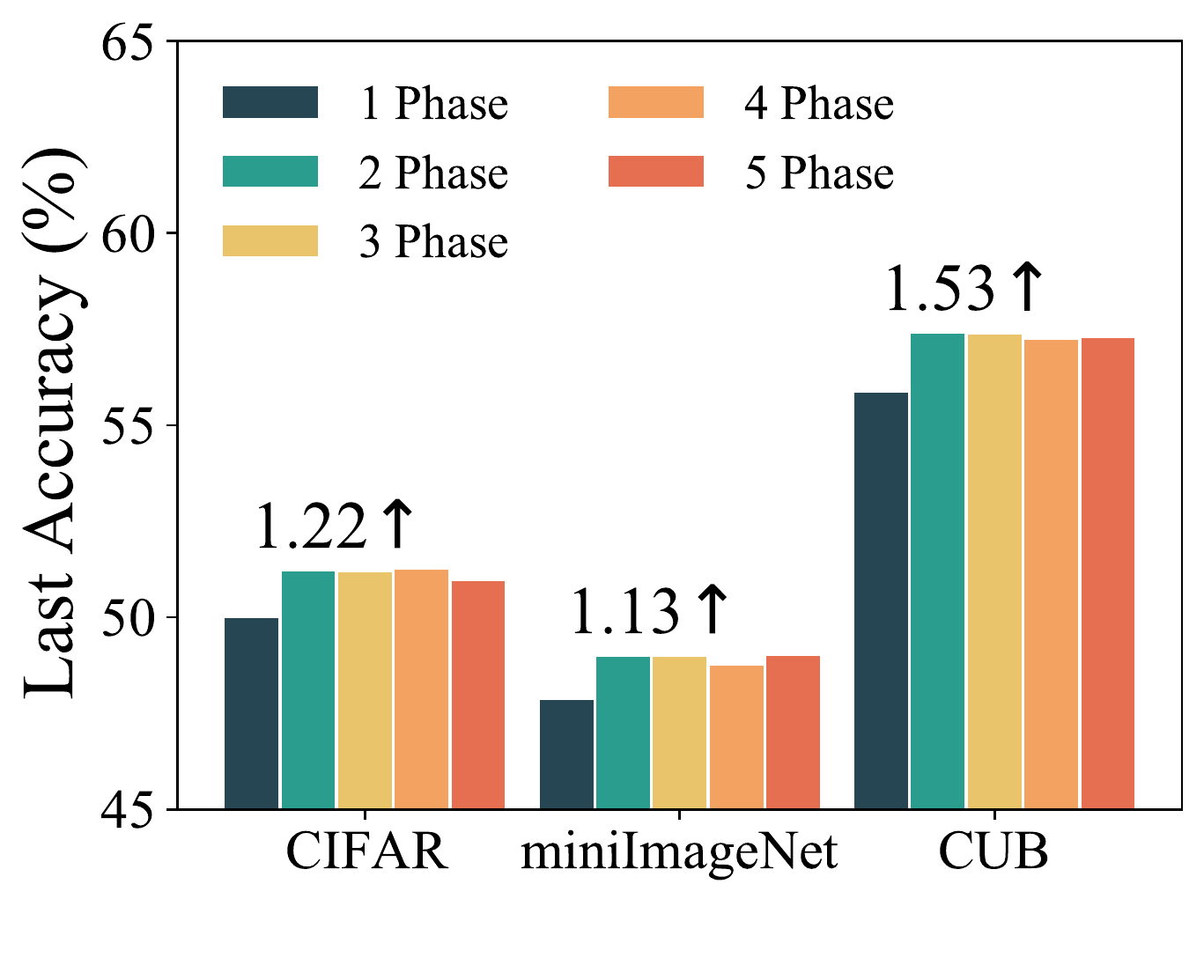}		\label{fig:meta-stage}}
		\hfill
		\subfigure[Multiple trials]
		{	\includegraphics[width=.64\columnwidth]{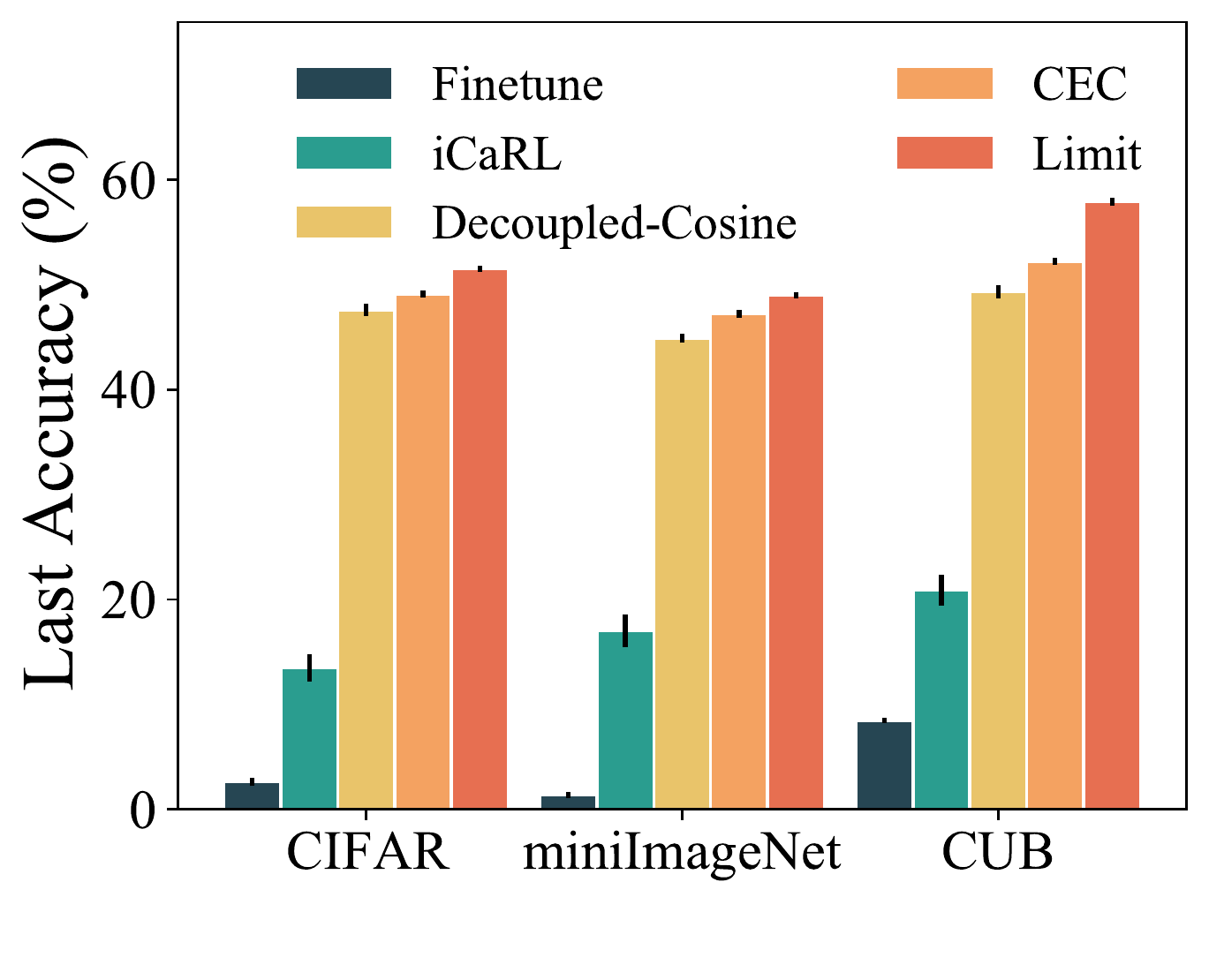}		\label{fig:multiple-run}}
		\\
		\subfigure[Incremental shot on CIFAR100]
		{	\includegraphics[width=.64\columnwidth]{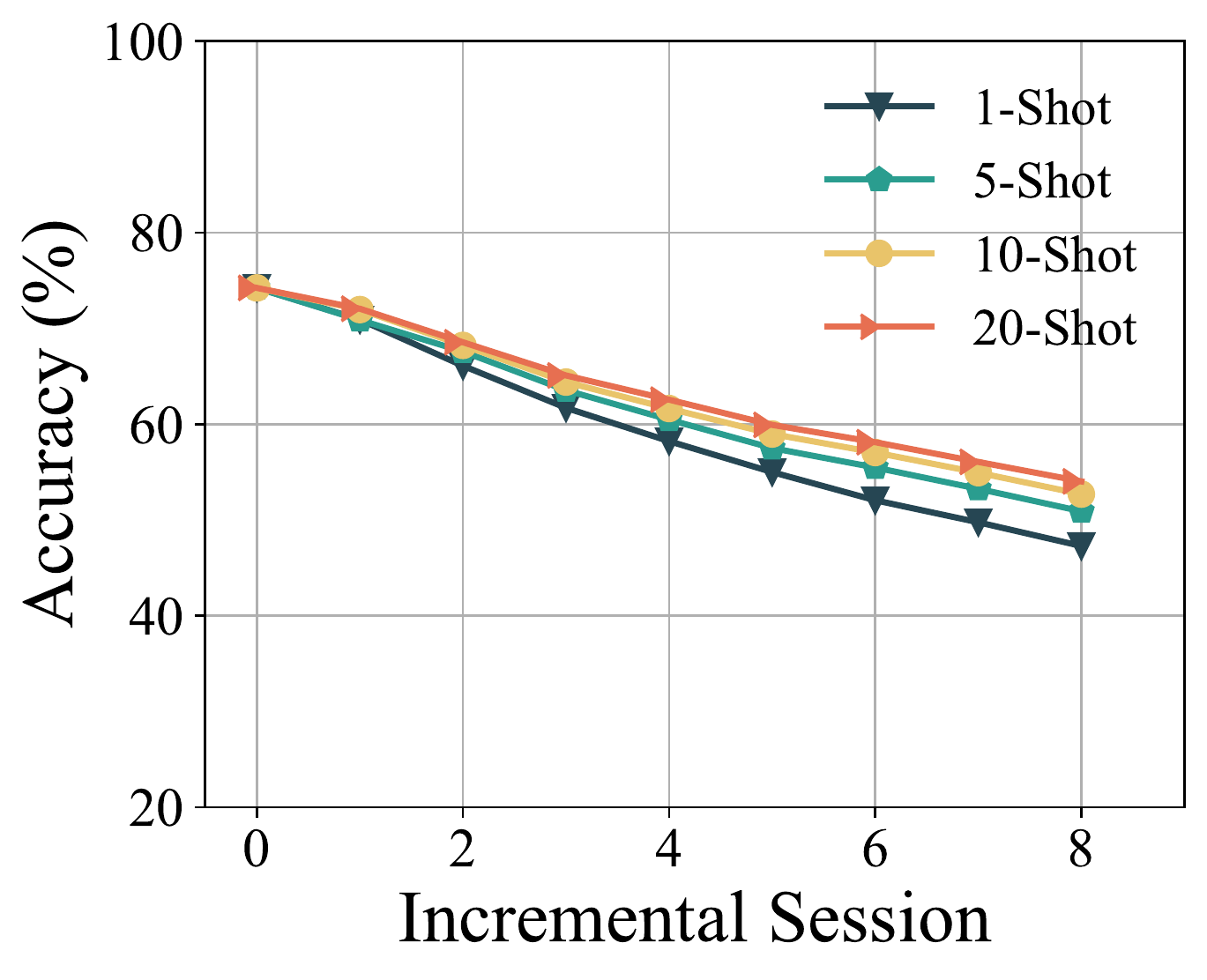}
			\label{fig:testshot-cifar}}
		\hfill
		\subfigure[Incremental shot on \textit{mini}ImageNet]
		{	\includegraphics[width=.64\columnwidth]{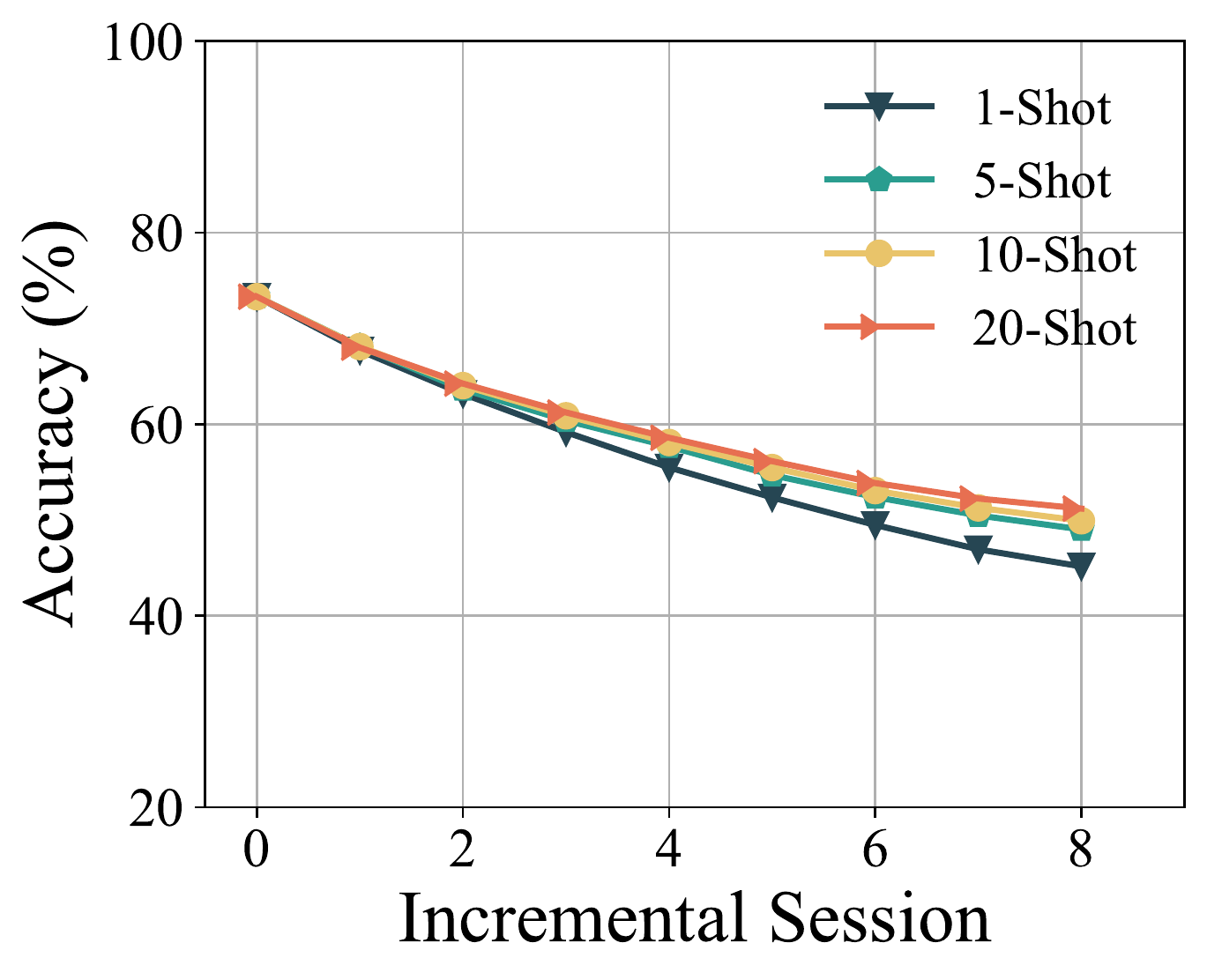}
			\label{fig:testshot-mini}}
		\hfill
		\subfigure[Incremental shot on CUB]
		{	\includegraphics[width=.64\columnwidth]{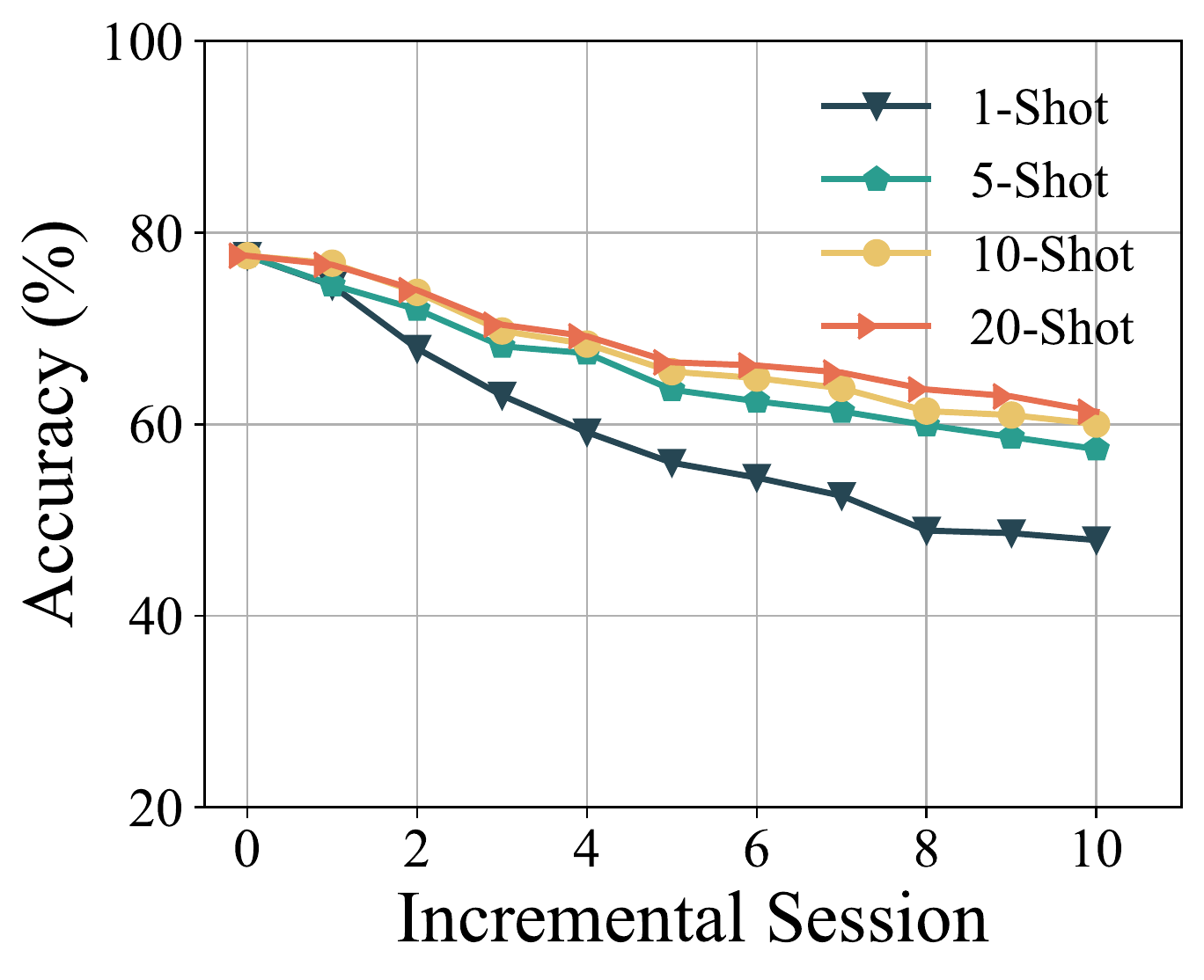}
			\label{fig:testshot-CUB}}
	\end{center}
	\caption{Analysis of the hyper-parameters. Large way and large shot in fake-incremental learning, more than one fake-incremental phase, and more shots of new classes will improve the performance.
	} \label{figure:ablation}
\end{figure*}

\experimentsection{Benchmark Comparison}
In this section, we report the experimental results on three benchmark datasets, \ie, CIFAR100, CUB200, and \textit{mini}ImageNet in  Figure~\ref{figure:benchmark}.
The detailed values are reported in Table~\ref{table:cub}~\ref{tab:cifar}~\ref{tab:mini} 
We report the baseline performance from~\cite{tao2020few,zhang2021few}. 
The results of \name are measured by training 2-phase fake-incremental tasks.

From Figure~\ref{figure:benchmark}, we can infer that \name consistently outperforms the SOTA methods on these datasets. For CIFAR100 and \textit{mini}ImageNet, \name outperforms the runner-up method by 1.5\%-2\%. The improvement on CUB200 is much greater, which reaches 5\%. Finetune does not consider regularizing former knowledge, which suffers catastrophic forgetting and gets the worst performance. Correspondingly, class-incremental learning algorithms consider maintaining former knowledge to resist forgetting. iCaRL restricts the old class discriminability by knowledge distillation, but it easily overfits on few-shot new classes and performs poorly in FSCIL. 
EEIL considers an extra balanced-finetuning process, which alleviates the forgetting to some extent. Rebalancing enhances the incremental model with contrastive training, but such a learning process is hindered by the limited instances. However, we can observe the overfitting phenomena of these CIL methods, indicating that CIL methods are not suitable for few-shot inputs. 
To this end, some algorithms are proposed for tailoring few-shot class-incremental learning scenarios. TOPIC utilizes a neural gas structure, which can maintain the topology of features between many-shot old classes and few-shot new classes. As a result, during the incremental learning process, the overfitting phenomenon is alleviated by such topology restriction, and TOPIC gets better performance than CIL methods. Motivated by this, CEC is proposed to decouple the learning process and further resist overfitting. The decoupled training paradigm generalizes the discriminability from base classes to new classes, obtaining the runner-up performance among all compared methods. Note that Decoupled-Cosine/DeepEMD and CEC also adopt the prototypical network protocol. \emph{The improvement of \name over these methods implies that our training paradigm is more proper for few-shot class-incremental learning}.

 When comparing \name to CEC, we can infer that \name consistently outperforms it in these benchmark settings. There are three reasons for such a performance gap: (1) We provide a novel sampling framework directly derived from the expected risk in Eq.~\ref{eq:fscil_risk}. Since FSCIL has multiple incremental sessions, \name directly optimizes the expected risk in Eq.~\ref{eq:meta-risk}, while CEC only minimizes the approximation with 1-stage tasks. (2) CEC uses image rotation to synthesize new classes, which is irrelevant to the FSCIL context. By contrast, we sample new classes from the base session, which is relevant and beneficial to real FSCIL tasks. (3) The meta-calibration module implemented with transformer can encode the calibration information between old and new classes.

We report the detailed value of these benchmark datasets in Table~\ref{table:cub}~\ref{tab:cifar}~\ref{tab:mini}, and show the corresponding performance dropping rate of each compared method. The last column shows the relative improvement, indicating how much our \name outperforms the other methods in terms of PD, \ie, resisting forgetting. 
We can infer from these tables that \name  consistently resists forgetting in the few-shot class-incremental tasks.

Apart from these benchmark datasets, we also conduct experiments on the popular large-scale datasets, \ie, ImageNet100 and ImageNet 1000. We report the incremental performance of \name and the competitive methods, \ie, CEC and Decoupled-Cosine in Figure~\ref{figure:imagenet}. As we can infer from these figures, \name consistently outperforms CEC in the large-scale learning scenario. To conclude, \name outperforms the current state-of-the-art method with a substantial margin on large-scale and small-scale datasets.

\subsection{Visualization of Confusion Matrix}

In this section, we visualize the confusion matrix after the last session for different methods on CUB200. The results are shown in Figure~\ref{figure:confmat}. The former 100 classes are base classes, and the rest 100 classes are incremental classes. 
Warm colors indicate higher accuracy in these figures, and cold colors indicate lower accuracy.

The confusion matrix of finetune is shown in Figure~\ref{fig:ft}, and we can infer that the method tends to predict the labels of the last session. Finetune easily falls overfitting on these new classes and suffers severe catastrophic forgetting. Figure~\ref{fig:icarl} shows the confusion matrix of iCaRL, which resists forgetting via knowledge distillation. iCaRL works better than finetune, with the diagonal brighter. However, it also tends to predict instances as new classes and suffers catastrophic forgetting. Figure~\ref{fig:cosine} shows the confusion matrix of Decoupled-cosine. It follows the prototypical network framework and utilizes a cosine classifier for classification. The results indicate that replacing classifiers with prototypes will not change the embedding, and overfitting phenomena will be alleviated. 
Although Decoupled-cosine maintains old class performance, the accuracy of new classes is not good enough. The results of \name are shown in Figure~\ref{fig:ours}, which has more competitive performance on \emph{new classes}.  The visualization of the confusion matrix indicates that \name adapts to new classes with a generalizable feature and stably resists catastrophic forgetting.

\begin{table}[t] 
	\caption{Accuracy analysis of base and incremental classes after the last incremental session on CUB200. \name improves the accuracy on new classes with a more generalizable feature embedding. }
	\centering
	{\begin{tabular}{lccc}
			\addlinespace
			\toprule
			{Methods} &{Base}  &{Incremental} &{Harmonic Mean}  \\
			\midrule
			Decoupled-Cosine & 71.5 &  28.8 &41.1\\
			CEC & 71.1 & 33.9 &45.9 \\
			\midrule
			\name & \bf 73.6 & \bf41.8 &\bf 53.3\\
			\bottomrule
		\end{tabular}\label{table:knownandunknown}}
\end{table}

To quantitatively measure the generalization ability of the model, we also report the average accuracy of base classes (classes in $Y_0$) and incremental classes (classes in $Y_1\cup\cdots Y_{B}$) after the last incremental session in Table~\ref{table:knownandunknown}.
We also follow~\cite{Cheraghian_2021_CVPR,ye2021learning} and report the harmonic mean between old and new classes.
 For comparison, we report the most competitive compared methods, \ie, Decoupled-Cosine and CEC. Table~\ref{table:knownandunknown} indicates that the performances for base classes are almost the same.
In contrast, CEC trains a graph model to incrementally update tasks and improves the accuracy of new classes by 5\%. However, \name considers sampling the multi-phase incremental sessions instead of single-phase. It utilizes the transformer architecture to extract invariant information for model extension, which achieves a remarkable improvement of 13\% on new classes.
Experimental results validate that \name learns a more generalizable feature embedding during the fake-incremental tasks.

\experimentsection{Analysis of Hyper-Parameters} \label{sec:hyperparam}
In this section, we study the influence of hyper-parameters on the final performance. 
In detail, we change the fake-shot/way in  fake-incremental learning, the number of fake phases, training instance selection, and test shot to find out their impact on final results.

\subsubsection{Fake-Way/Shot}
We first report the final accuracy on CIFAR100 by varying the fake-way and fake-shot in Figure~\ref{fig:cifar-wayshot}.
According to the discussions in Section~\ref{sec:discussions}, \name does not require the meta-training and testing stage to follow the `same-way same-shot' protocol. As a result, we fix the other settings the same as in benchmark experiments and change the fake-incremental way from $\{1, 5, 10, 15, 20\}$. We also choose the fake-incremental shot from $\{1, 5, 10, 15, 20\}$, resulting in 25 compared results. We can infer from Figure~\ref{fig:cifar-wayshot} that \name prefers large fake-training way and fake-training shot, \ie, training with 20-way 20-shot gets the best performance. Nevertheless, we also notice the influence of fake-incremental way is stronger than fake-incremental shot.

\subsubsection{Fake-Incremental Phase}
We then report the final accuracy on three benchmark datasets by varying the number of fake-incremental phases $C$. 
We sample a 5-way 5-shot fake-incremental task for each phase and choose $C$ from $\{1, 2, 3, 4, 5\}$. 
From Figure~\ref{fig:meta-stage}, we can infer that the performance of more than one incremental phase is better than that of one phase, which verifies the effectiveness of our training paradigm for FSCIL.
The number above each cluster indicates the improvement of multi-phase training over single-phase training.
However, we also find the improvement is trivial for more than two phases, and we set the sampling phases to 2 for all datasets.

\subsubsection{Multiple Trials}
The current benchmark-setting is defined in~\cite{tao2020few}, where the methods are evaluated with the same training instances for \emph{only one time}. To empirically evaluate the robustness of algorithms, we conduct more trials and report the average and standard deviation.
In detail, following the same class split as~\cite{tao2020few}, we suggest sampling different few-shot instances as the training sets $\{\D^1,\D^2,\cdots,\D^B\}$. Specifically, we first filter out the instances of each class and then randomly sort them, picking the first $K$ instances. The sampling process can be reproduced by assigning specific random seeds.\footnote{In our implementation, we use random seeds from $\{1,2,\cdots, E\}$, where $E$ stands for the total rounds.} 
We sample different FSCIL episodes 30 times and report the average final accuracy and standard deviation. The results of five typical methods are shown in Figure~\ref{fig:multiple-run}.

We can infer from Figure~\ref{fig:multiple-run} that \name works robustly facing different episode combinations. Besides, iCaRL utilizes knowledge distillation to prevent forgetting, and its performance varies between different task combinations. The ranking order of these five methods is the same as reported in the benchmark tables, indicating \name consistently outperforms other methods.

\subsubsection{Incremental Shot}
In the FSCIL setting, each incremental training set $\mathcal{D}^b$ can be formulated as $N$-way-$K$-shot.
We also change the shot number $K$  of each new class during the incremental tasks, and report the learning trends of \name in Figure~\ref{fig:testshot-cifar}, ~\ref{fig:testshot-mini}, and~\ref{fig:testshot-CUB}. 
 We keep the testing way the same as the benchmark-setting and vary the shot number from $\{1, 5, 10,  20\}$.
Results indicate that the model receives more information for new classes with more instances from each class. 
 Hence, the estimation of prototypes will be more precise, and the prediction performance will correspondingly improve. However, the increasing trend will converge as more training instances are available, \eg, $K=20$.

\begin{figure}[t]
	\begin{center}
		\subfigure[Five old classes]
		{	\includegraphics[width=.465\columnwidth]{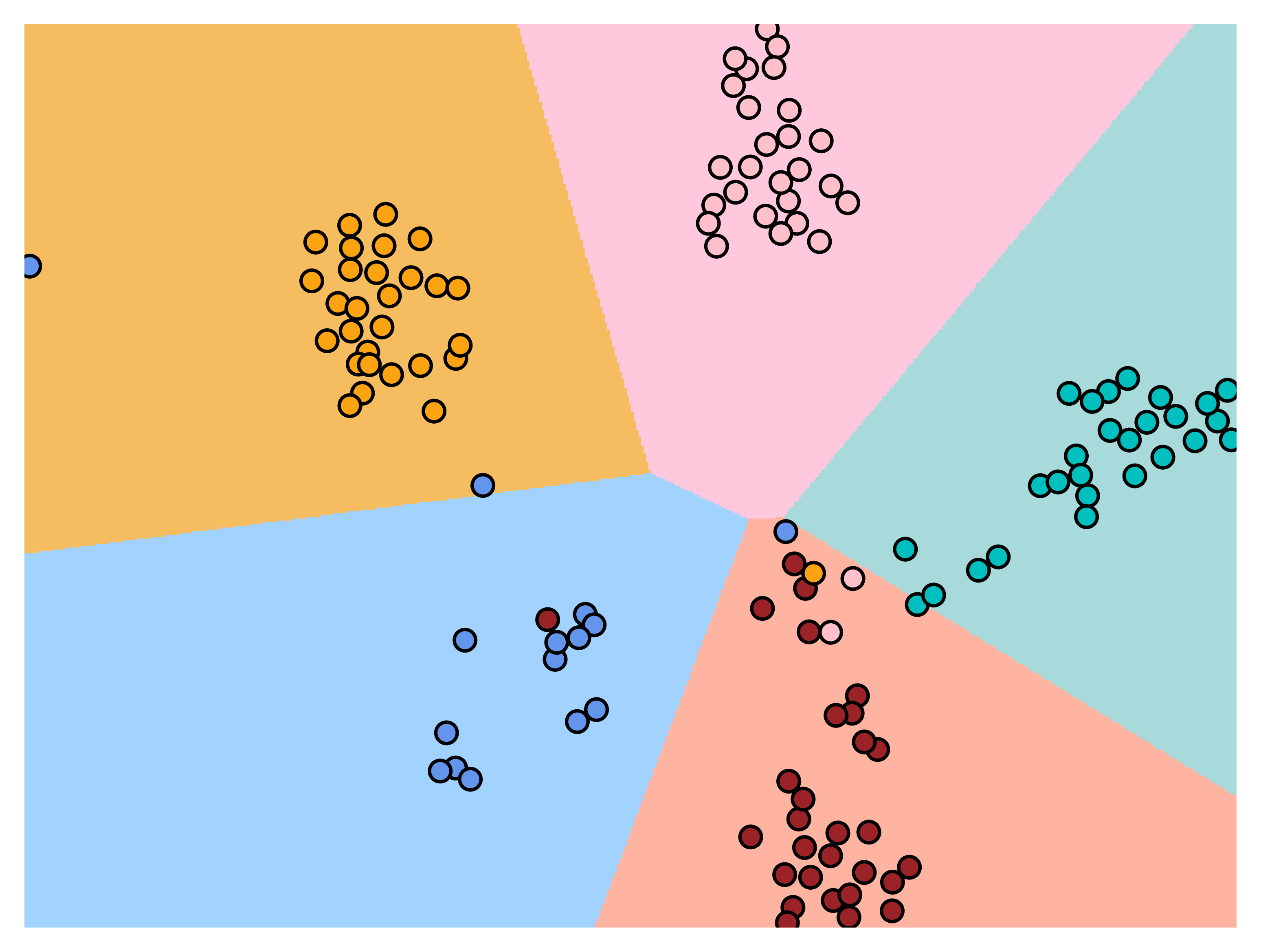} \label{fig:tsne1}}
		\subfigure[Five old \& five new classes]
		{		\includegraphics[width=.465\columnwidth]{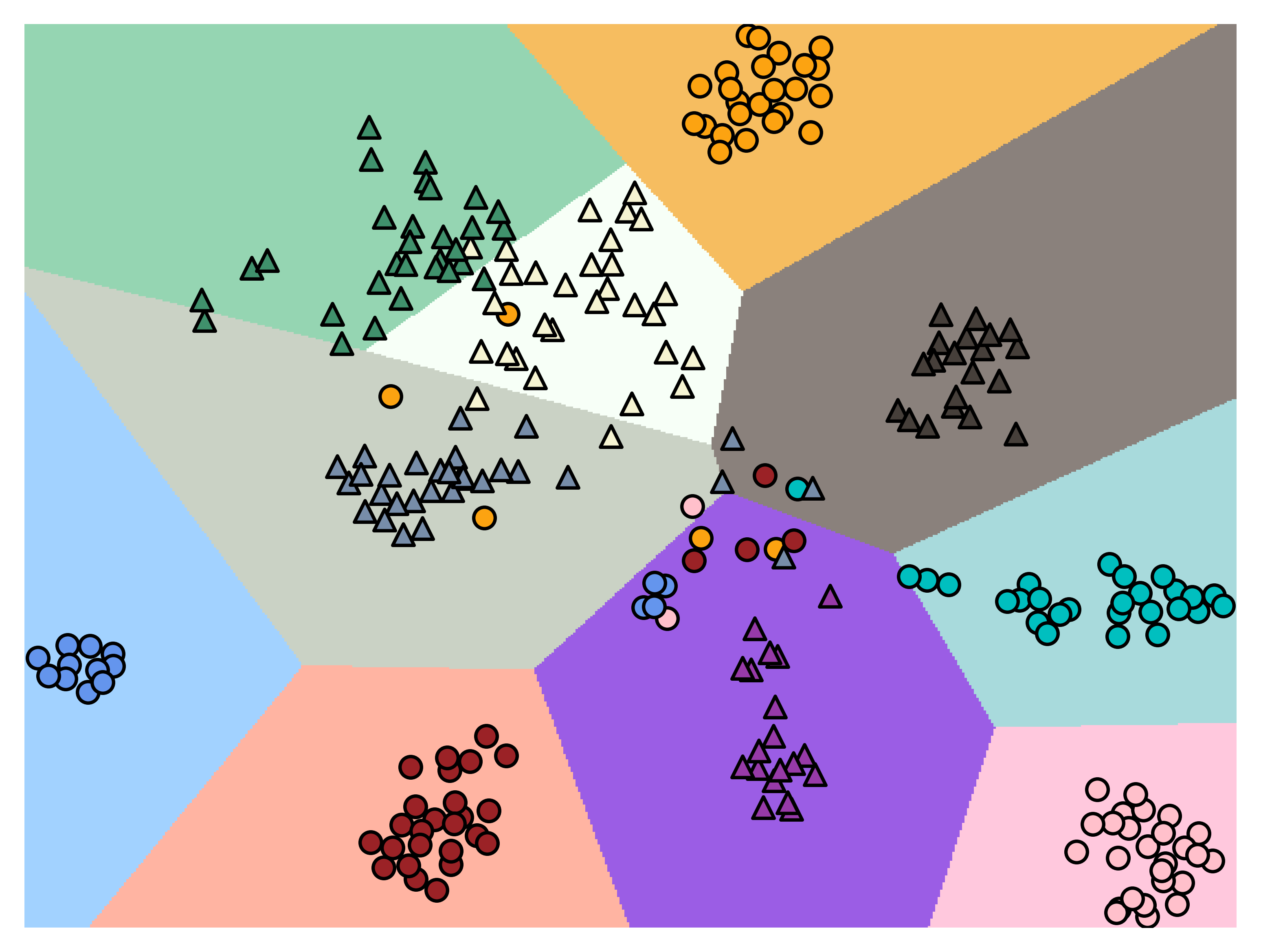} \label{fig:tsne2}}
		
	\end{center}
	\caption{ t-SNE visualization of the learned decision boundary on CUB200 between two sessions. Old classes are shown in dots, and new classes are shown in triangles. The shadow region represents the decision boundary of each class. 
	} \label{figure:tsne}
\end{figure}

\experimentsection{Visualization of Decision Boundaries}
In this section, we visualize the learned decision boundaries on the CUB200 dataset.  We use t-SNE~\cite{van2008visualizing} to visualize the test instances and corresponding decision boundaries of each class in 2D, as shown in Figure~\ref{figure:tsne}. 

In Figure~\ref{fig:tsne1}, we show model before incremental updating. 
Shadow regions represent the decision boundary of five base classes, and embeddings of test instances are shown with dots.
After that, we extend our model with $5$-way $5$-shot new classes and visualize the updated decision boundary in Figure~\ref{fig:tsne2}. We can infer that the meta-calibration module helps to adapt the prototype and calibrate the decision boundary between old and new classes. 
As a result,  \name works competitively with few-shot inputs, which efficiently calibrates old and new classes in the few-shot class-incremental setting. The figures also indicate that incorporating  the knowledge of new classes does not harm the classification performance of old classes, \ie, \name maintains classification performance and resists catastrophic forgetting.

\experimentsection{Visualization of Meta-Calibration Module}

\begin{figure}[t]
	\begin{center}

		{	
			\includegraphics[width=.95\columnwidth]{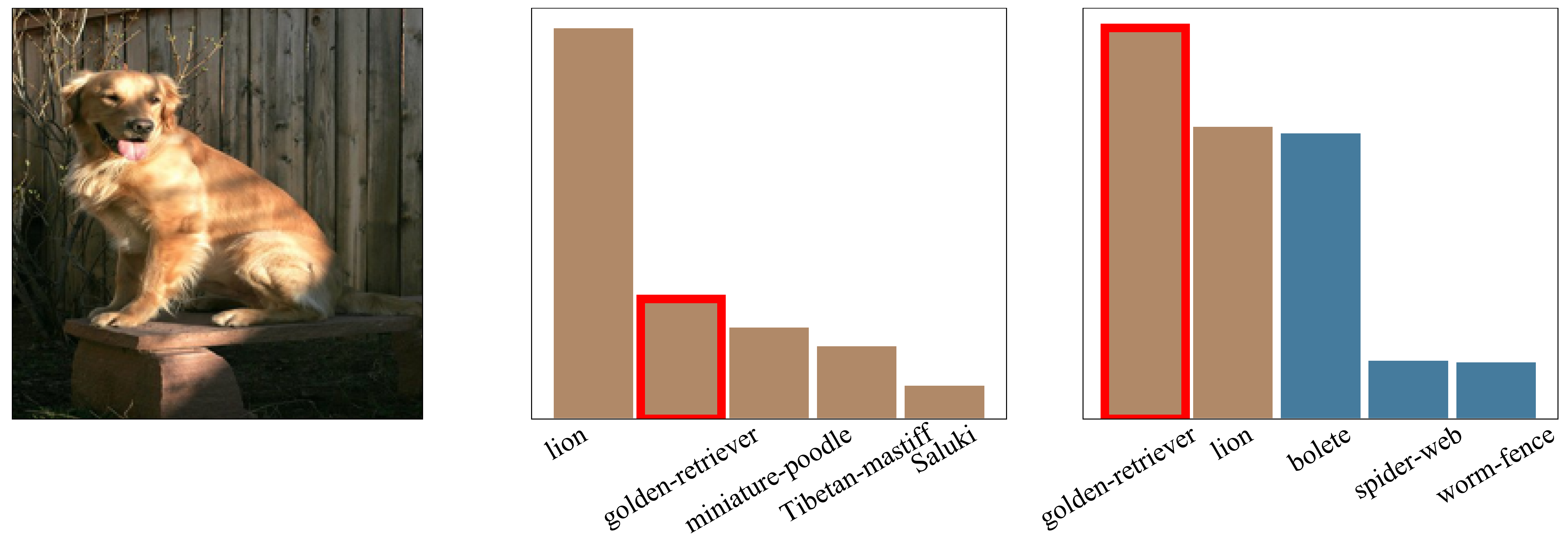}
			\includegraphics[width=.95\columnwidth]{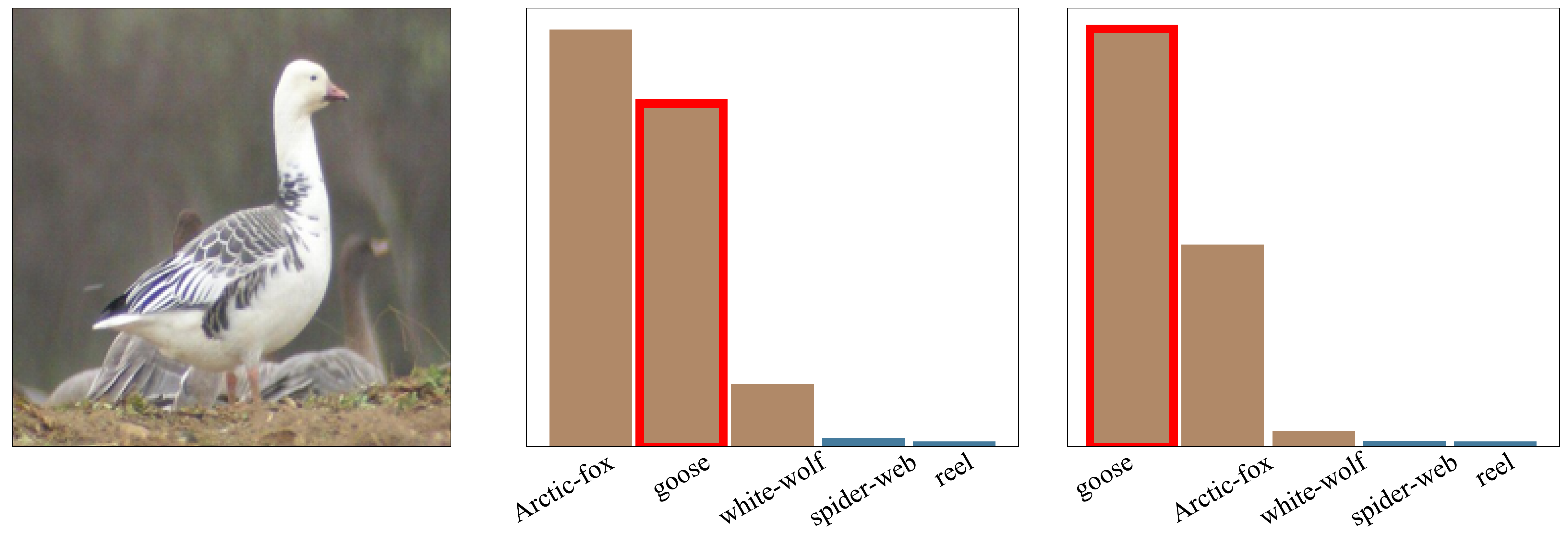}
			\includegraphics[width=.95\columnwidth]{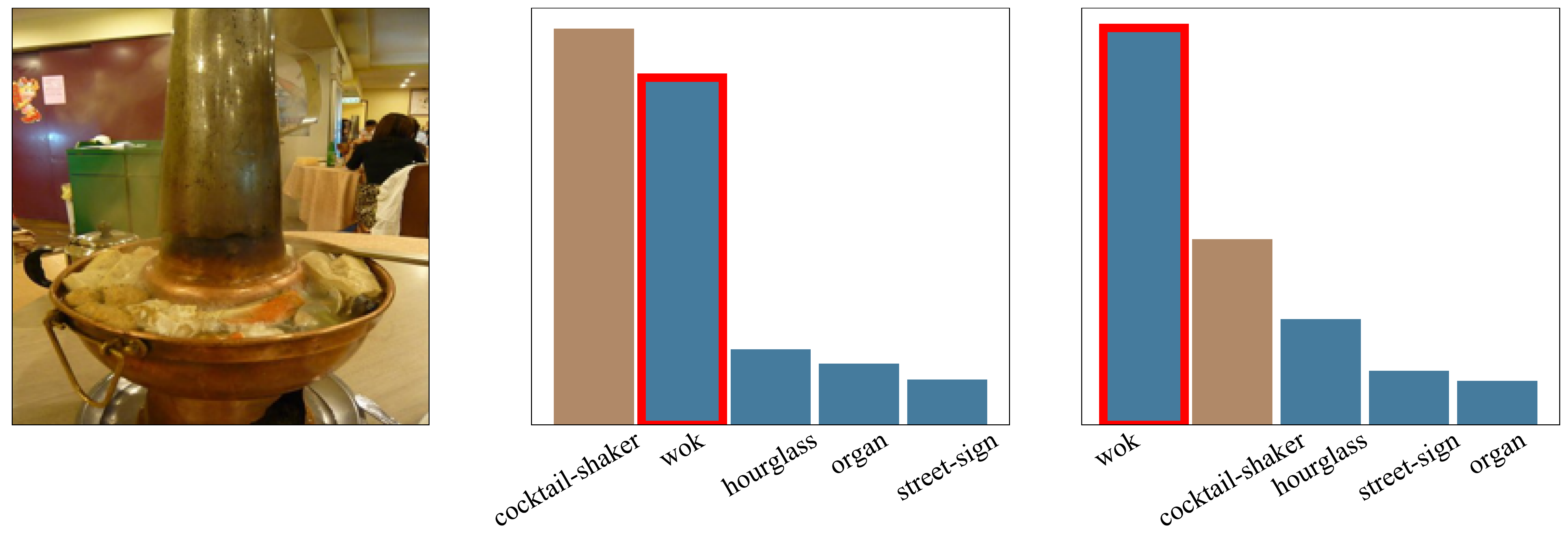}
			\includegraphics[width=.95\columnwidth]{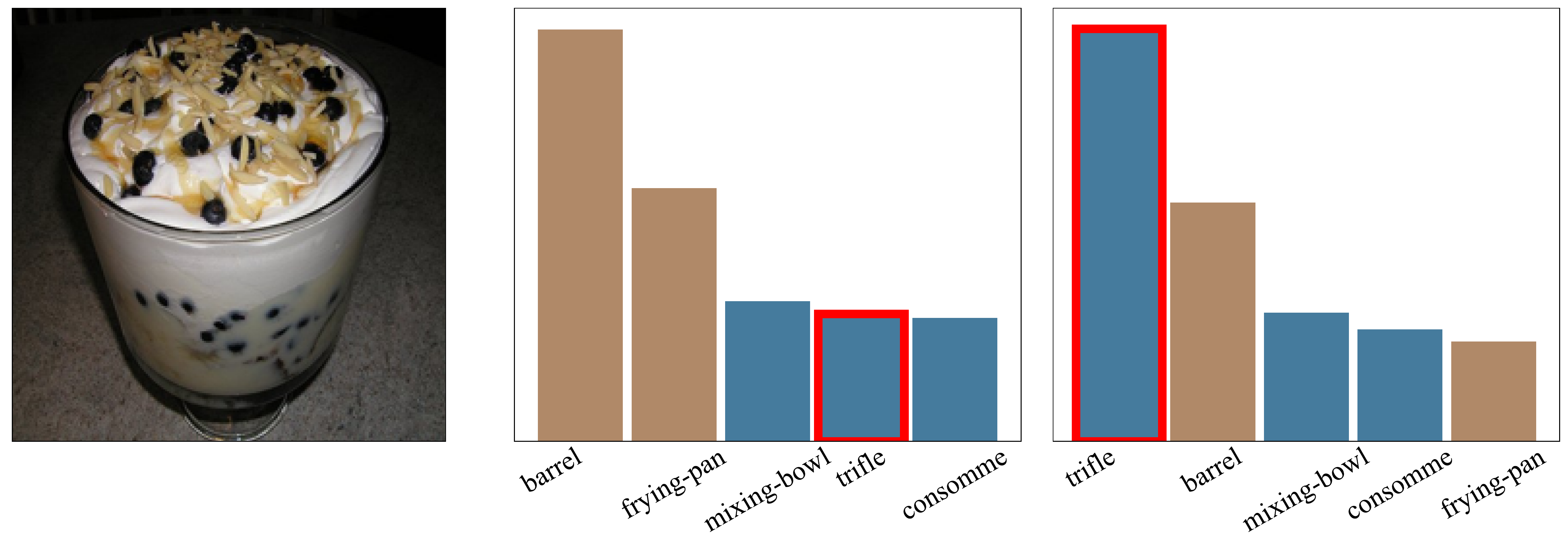}

		}
		
	\end{center}
	\caption{ Visualization of the prediction probability before and after meta-calibration on \textit{mini}ImageNet. The first row indicates original images. The second row indicates the top-5 output probability before meta-calibration. The third row indicates the top-5 output probability after meta-calibration.
		Base classes are shown with brown color, and incremental classes are shown in blue. The ground-truth class is shown with red edges.
	} \label{figure:meta-calibration}
\end{figure}

In this section, we visualize the prediction results before and after meta-calibration and analyze their differences. We choose images from \textit{mini}ImageNet and use the model trained under the benchmark setting. The results are shown in Figure~\ref{figure:meta-calibration}. We show the original images in the first row, the top-5 prediction probability before meta-calibration module (\ie, $W^\top \phi(\x)$) in the second row, and the top-5 prediction probability after meta-calibration (\ie, ${\tilde{W}}^\top \tilde{\phi}(\x)$) in the last row.

The predicted probabilities of base classes are shown in brown bars, and the predicted probabilities of incremental classes are shown in  blue bars.
 The probability of the ground-truth label is denoted with red edges. 
 As shown in these figures, the top two figures are from the base classes, and the bottom two are from the incremental classes.
  For the many-shot base classes, the model may have wrong predictions, \ie, predicting a golden retriever into a lion or predicting a goose into an arctic fox. To this end, the meta-calibration module can correct the model and increase the probability of the ground-truth class with the transformer. 
 
 When switching to the inference of new classes, since the model has only seen few-shot training images, it tends to predict them as base classes, \eg, predicting a wok into a cocktail-shaker or predicting a trifle into a barrel.
  Under such circumstances, the meta-calibration module will help to re-rank the ordering of these outputs and increase the output probability on new classes. These visualizations indicate that the meta-learned calibration module has encoded the inductive bias in the fake-incremental learning process and generalized it into the inference time. These conclusions are consistent with Table~\ref{table:knownandunknown} and Figure~\ref{figure:confmat} that \name improves the learning ability of new classes.

\experimentsection{Analysis of Incremental Task Sampling,  Model Efficiency and Data Augmentations}\label{sec:c2cec}

In this section, we analyze the task sampling strategy and model efficiency of different methods. In detail, we interchangeably use the fake task sampling strategy to find out the proper one. Besides, we report the training time and model parameters for different methods to get a holistic evaluation. We also analyze the results of different methods with different augmentations.

\begin{table}[t]
	\centering
	\caption{Ablation study of fake task sampling strategy. Fake incremental tasks (FIT) are our sampling strategy, and pseudo incremental learning (PIL) are from CEC. We interchangeably use these task sampling strategy to train CEC and \name, and report the final accuracy on CUB200 and CIFAR100. }
	\resizebox{0.8\columnwidth}{!}{
		\begin{tabular}{l|cc|cc}
			\addlinespace
			\toprule
			Dataset& \multicolumn{2}{c|}{CUB200} & \multicolumn{2}{c}{CIFAR100}\\
			Task Sampling& FIT & PIL & FIT & PIL \\
			\midrule
			{CEC} & \bf 53.01 & 52.28 &\bf 49.60 & 49.14\\
			{\name} & \bf 57.41 & 56.37 & \bf 51.23 & 51.03\\
			\bottomrule
		\end{tabular}
	}
	\label{tab:ablation_fake_task}
\end{table}

\subsubsection{Task Sampling Strategy}
According to the discussions in Section~\ref{sec:discussions}, there are other ways to sample fake-incremental tasks for meta-training. We choose the most typical method, \ie, CEC, to conduct the ablation analysis. Specifically, we interchangeably use the fake task sampling process in CEC and our \name to train the model and report the last accuracy of these methods. In CEC, the meta-learning tasks are denoted as pseudo incremental learning (PIL), which is called fake-incremental tasks (FIT) in ours. The results are reported in Table~\ref{tab:ablation_fake_task}. We can infer from Table~\ref{tab:ablation_fake_task} that our fake task sampling strategy can improve the performance of CEC, which verifies the strengths discussed in Section~\ref{sec:discussions}. Besides, when changing our fake-task sampling strategy to the way of CEC, the performance of \name degrades. These conclusions imply that our sampling strategy is a better choice for few-shot class-incremental learning. 

\subsubsection{Model Efficiency}
\begin{figure}[t]
	\begin{center}
		\subfigure[Training time]
		{\includegraphics[width=.48\columnwidth]{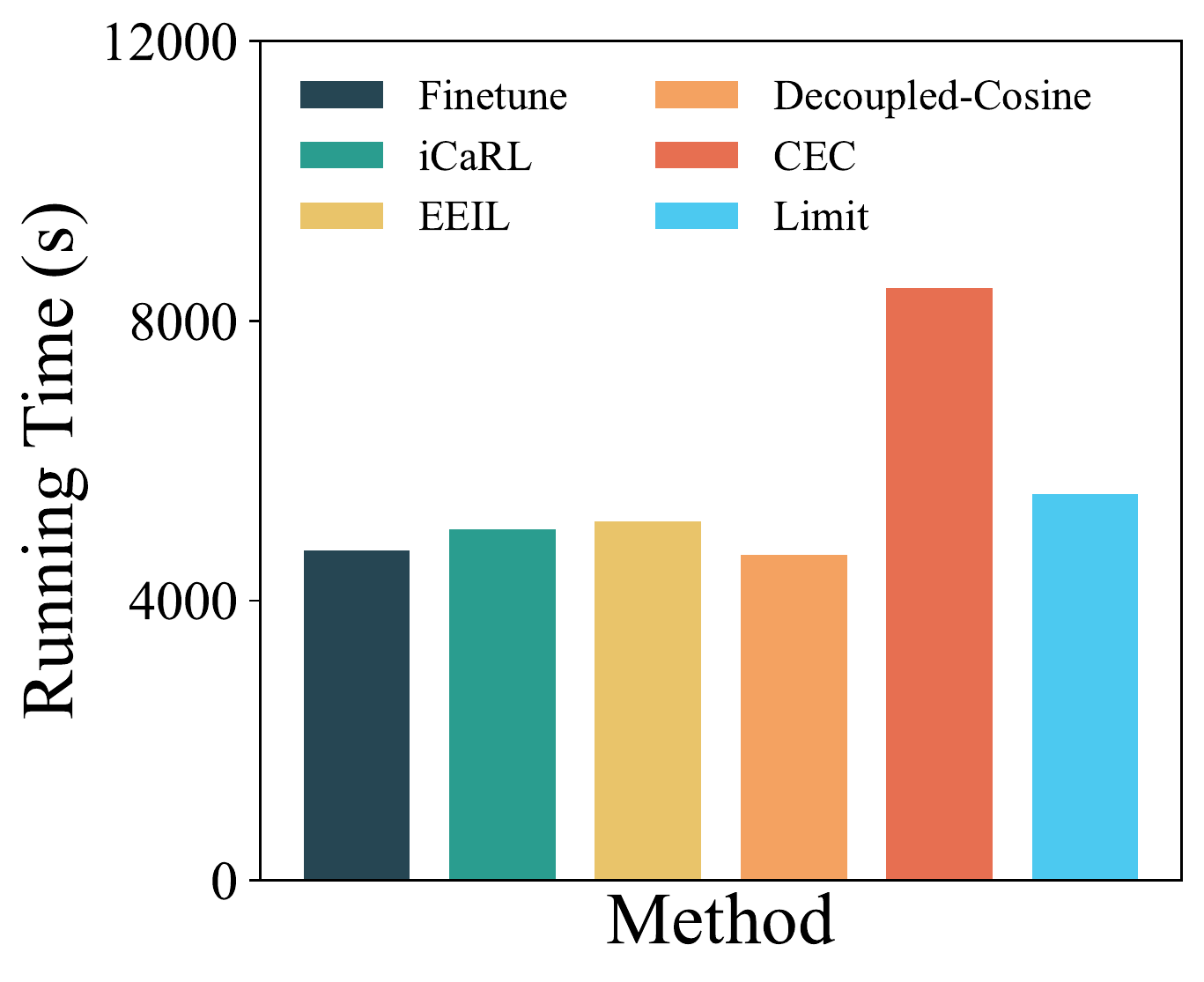}
			\label{figure:time_and_param_a}}
		\subfigure[Model parameters]
		{\includegraphics[width=.48\columnwidth]{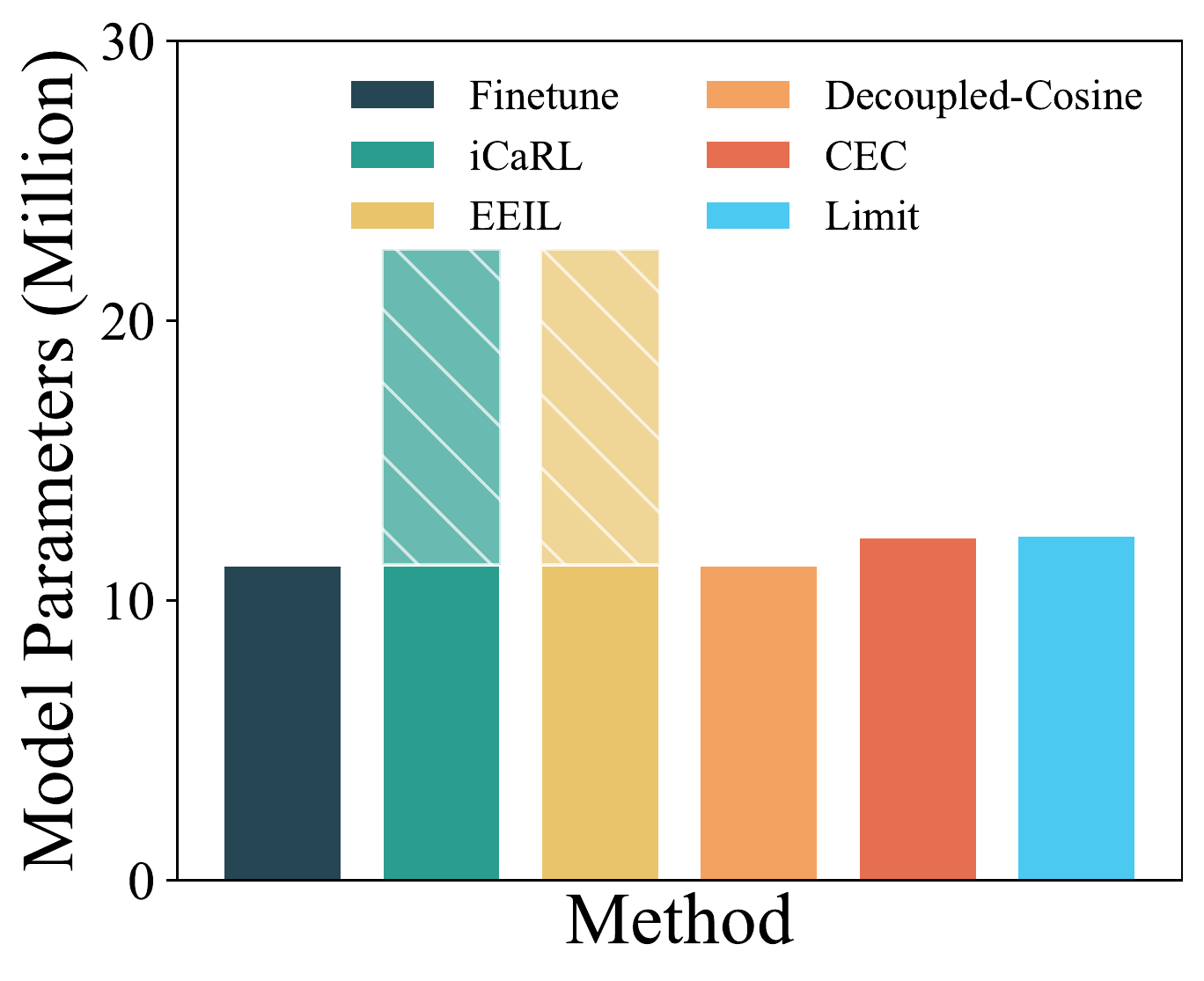}
			\label{figure:time_and_param_b}}
	\end{center}
	\caption{Running time and model parameter comparison of several typical methods.  The bars with shadow denote the parameters used during training but dropped during inference.
	} \label{figure:time_and_param}
\end{figure}

Since \name includes an extra transformer model to calibrate prototype and query instances, we report the model size and running time comparison of several typical methods on CUB200 in Figure~\ref{figure:time_and_param}.

The training time of different methods is reported in Figure~\ref{figure:time_and_param_a}. The training time of \name includes the pre-training process (which is the same for all methods) and the meta-training process.
We can infer that \name only requires slight extra training time to learn the meta-calibration module. The extra running time of \name is humble compared to that of CEC, and our running time is at the same scale as other methods. Besides, we can infer from Figure~\ref	{figure:time_and_param_b} that the extra model size of \name is at the same scale as CEC. The number of model parameters of the transformer is about 1 million, which is neglectable compared to the backbone network (ResNet18). There are some methods utilizing knowledge distillation to resist forgetting, \eg, iCaRL and EEIL. These methods need to save an extra backbone as the old model to provide supervision of knowledge distillation, and \name is more memory efficient during the training process.

\subsubsection{Data Augmentations} \label{sec:data_aug}

\begin{figure}[t]
	\begin{center}
		\subfigure[CUB200]
		{\includegraphics[width=.48\columnwidth]{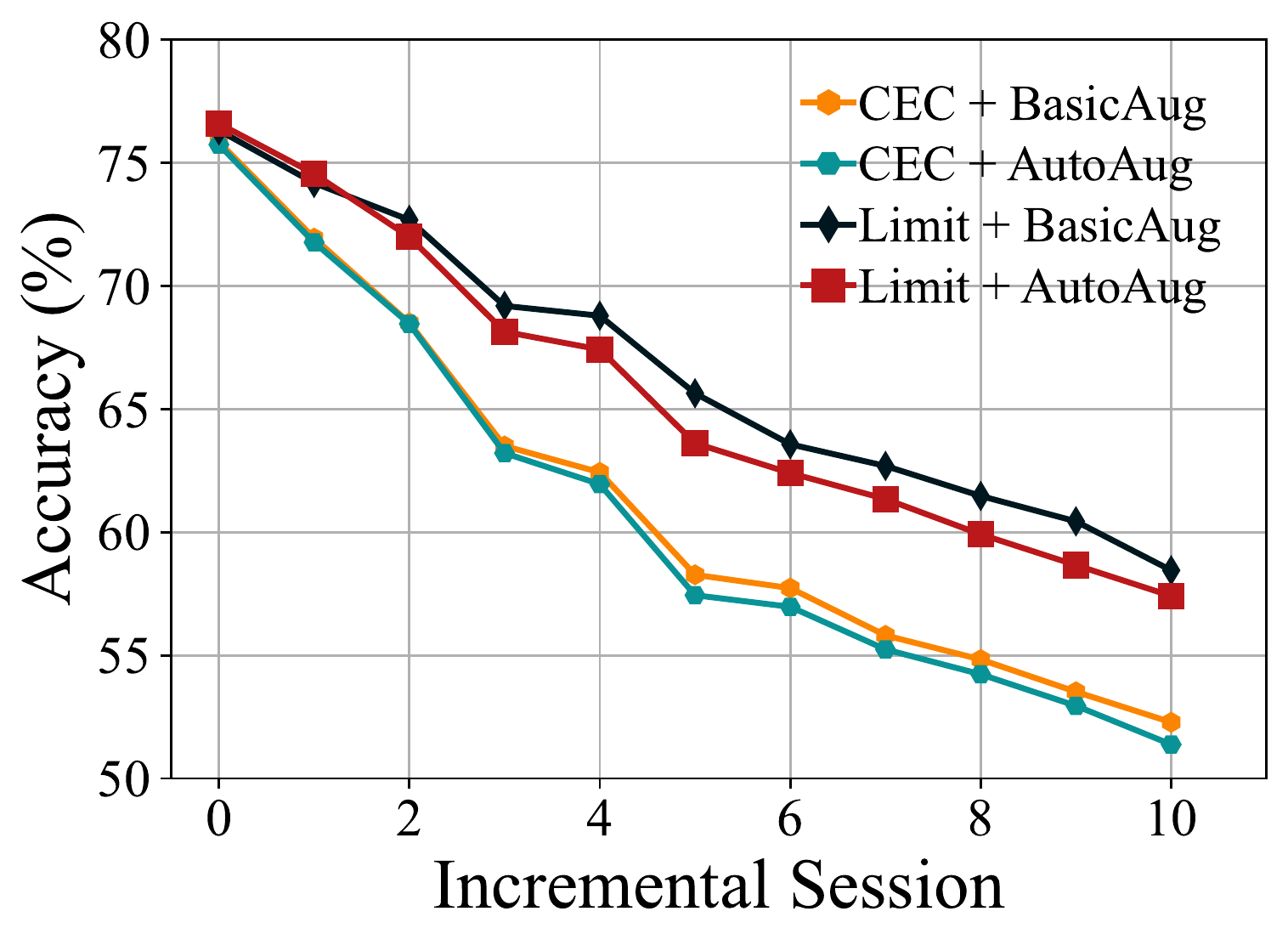}
			\label{figure:augmentation_a}}
		\subfigure[{\it mini}ImageNet]
		{\includegraphics[width=.48\columnwidth]{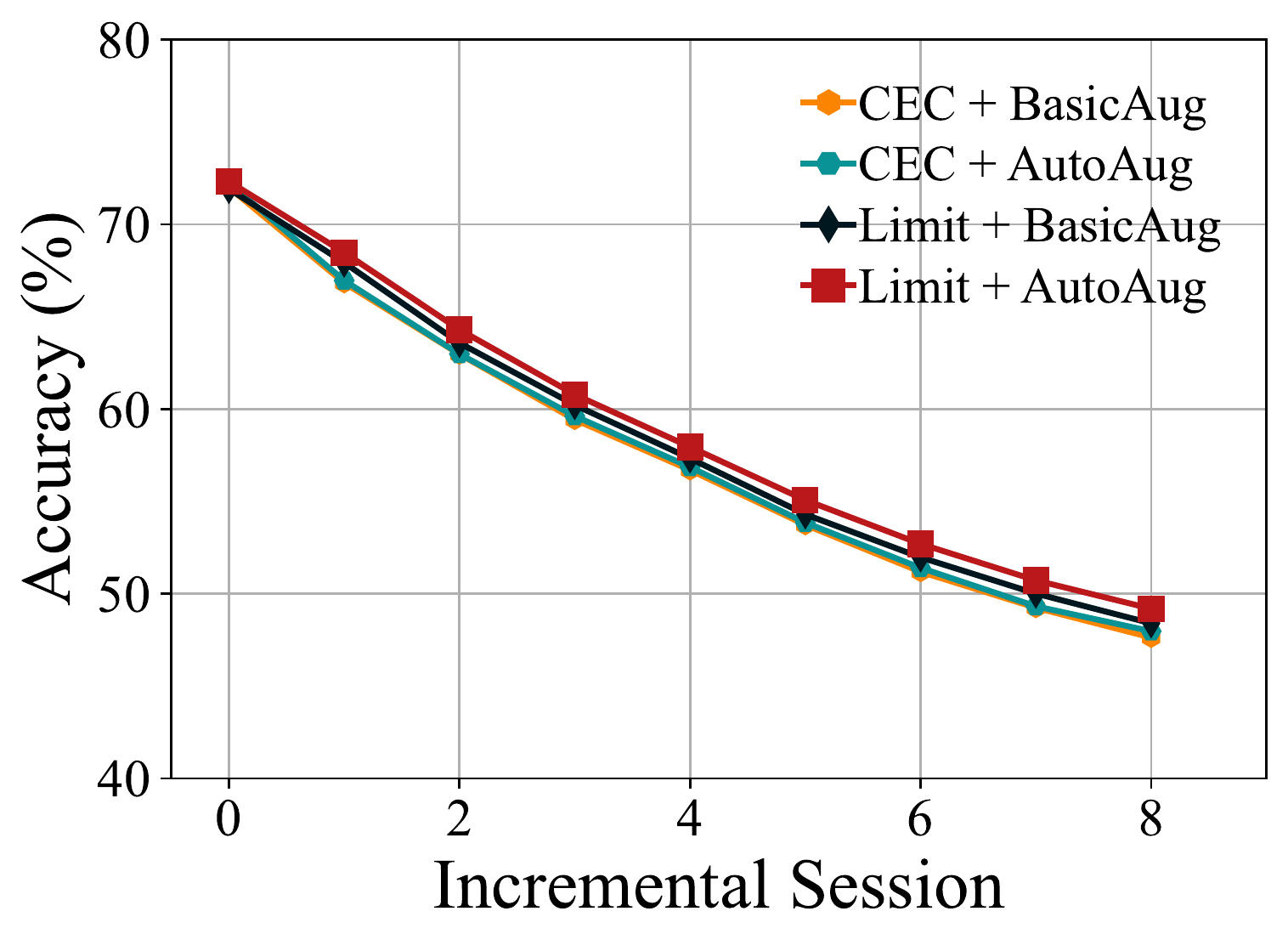}
			\label{figure:augmentation_b}}
	\end{center}
	\caption{Ablation study of data augmentations. Our proposed method outperforms CEC no matter with or without AutoAug.
	} \label{figure:augmentation_ablation}
\end{figure}

As a common technique in deep learning, augmentations are proven to be effective in model training. To investigate the influence of using these augmentations, we conduct experiments by combining different augmentations to ours and CEC, and report the results on benchmark datasets in Table~\ref{tab:ablation_augmentation}. We also plot the incremental performance in Figure~\ref{figure:augmentation_ablation}. Denote the augmentations in CEC as BasicAug (\ie, random crop and random horizontal flip). We train CEC and \name separately with BasicAug/AutoAug, resulting in four combinations. We can draw several conclusions from these ablations:

	\begin{table}[t]
	\centering
	\caption{Ablation study of augmentation strategies. BasicAug denotes the augmentation policy adopted in CEC. We interchangeably use these augmentation strategies to train CEC and \mame, and report the final accuracy. }
		\begin{tabular}{cccc}
		\addlinespace
		\toprule
		Dataset& {CIFAR100} &{CUB200} & {\textit{mini}ImageNet}\\
		\midrule
		{CEC + BasicAug} &49.14 & 52.28 & 47.63 \\
		{\name + BasicAug} &\bf 49.92 & \bf 58.45 & \bf 48.40 \\
		\midrule
		{CEC + AutoAug} & 49.55& 51.38 & 47.97 \\
		{\name + AutoAug} & \bf 51.23&  \bf 57.41 & \bf 49.19 \\
		\bottomrule
	\end{tabular}
	\label{tab:ablation_augmentation}
\end{table}

	\begin{itemize}
	\item Stronger augmentations can boost the performance of FSCIL models. Both CEC and \name facilitate from AutoAug and obtain better performance. CUB200 is the particular case where AutoAug leads to inferior performance. The main reason is the domain gap between CUB200 and ImageNet, since the augmentation policy is optimized for ImageNet.
	\item Our proposed method consistently outperforms CEC on both conditions, \ie, with or without AutoAug. Besides, comparing \name + BasicAug to CEC + AutoAug, we find \name can obtain competitive results against CEC even with weaker augmentations.	
\item Our proposed method facilitates more from the augmentations, \ie, the improvement of \name is larger than CEC. 
The main reason is that CEC relies on image rotation to synthesize new classes, which may conflict with the augmentation policies defined in AutoAug.
It indicates that our proposed method can be orthogonally combined with other useful tricks to further improve the incremental performance.
\end{itemize}

\section{Conclusion}

Real-world applications often face incremental datasets, and a model should learn new classes without forgetting old ones. Few-shot class-incremental learning is a challenging scenario, where overfitting and catastrophic forgetting co-occur. In this paper, we propose \name to simulate fake FSCIL tasks and prepare the model for future FSCIL tasks. In \mame, a new learning paradigm is proposed where we sample fake-incremental tasks and obtain generalizable features from diverse fake tasks. We also propose to encode the inductive bias into the meta-calibration module, which helps to calibrate between classifiers and the few-shot prototypes. 
The meta-calibration module also helps to generate instance-specific embedding and further improves the  performance. 
\name efficiently adapts to new classes and preserves old knowledge when learning new ones, consistently achieving state-of-the-art performance. 
Exploring other format set-to-set and calibration functions are interesting future works.

\ifCLASSOPTIONcompsoc
\section*{Acknowledgments}
\else
\section*{Acknowledgment}
\fi

This research was supported by National Key
R\&D Program of China (2020AAA0109401), NSFC (61773198, 61921006, 62006112), NSFC-NRF Joint Research Project under Grant 61861146001, Collaborative Innovation Center of Novel Software Technology and Industrialization, NSF of Jiangsu Province (BK20200313), CCF-Hikvision Open Fund (20210005).

\bibliographystyle{unsrt}
\bibliography{fscil}


%



\ifCLASSOPTIONcaptionsoff
  \newpage
\fi

\appendices

\section{Using Transformer as Set-To-Set Function}
As we mentioned in the main paper, we employ self-attention mechanism~\cite{lin2017structured,vaswani2017attention} to act as the meta-calibration module. 
Transformer is a store of triplets
in the form of (query $\mathcal{Q}$, key $\mathcal{K}$, and value $\mathcal{V}$). Elements in the query set are the ones we want to do the transformation. The transformer first matches a query point with each of the keys by computing the “query” – “key” similarities. Then the proximity of the key to the query point is used to weigh the corresponding values of each key. The transformed input acts as a residual value that will be added to the input.

\noindent\textbf{Basic Transformer:} Following the definitions in~\cite{vaswani2017attention}, we use $\mathcal{Q}, \mathcal{K}$, and $\mathcal{V}$ to denote the set of the query, keys, and values, respectively. All these sets are implemented by different combinations of task instances.
To increase the flexibility of the transformer, three sets of linear projections $\left(W_{Q} \in \mathbb{R}^{d \times d^{\prime}}, W_{K} \in \mathbb{R}^{d \times d^{\prime}}\right.$, and $\left.W_{V} \in \mathbb{R}^{d \times d^{\prime}}\right)$ are defined,\footnote{We omit the bias term for simplification.} one for each set. The points in sets are first projected by the corresponding projections:
\begin{align}
	\notag
	Q&=W_{Q}^{\top}\left[{\mathbf{x}_{q}} ; \quad \forall \mathbf{x}_{q} \in \mathcal{Q}\right] \in \mathbb{R}^{d^{\prime} \times|\mathcal{Q}|} \\
	K&=W_{K}^{\top}\left[{\mathbf{x}_{k}} ; \quad \forall \mathbf{x}_{k} \in \mathcal{K}\right] \in \mathbb{R}^{d^{\prime} \times|\mathcal{K}|} \\ \notag
	V&=W_{V}^{\top}\left[{\mathbf{x}_{v}} ; \quad \forall \mathbf{x}_{v} \in \mathcal{V}\right] \in \mathbb{R}^{d^{\prime} \times|\mathcal{V}|}
\end{align}
$|\mathcal{Q}|,|\mathcal{K}|$, and $|\mathcal{V}|$ are the number of elements in the sets $\mathcal{Q}, \mathcal{K}$, and $\mathcal{V}$, respectively. Since there is a one-to-one correspondence between elements in $\mathcal{K}$ and $\mathcal{V}$ we have $|\mathcal{K}|=|\mathcal{V}|$.

The similarity between a query point\footnote{In our implementation, the query instance can be from classifier and query embedding, \ie, $\mathcal{Q}=\mathcal{K}=\mathcal{V}=[\hat{W}, \phi(\x)]$. Without loss of generality, here we use $\mathbf{x}_{q}$ to denote a query instance in $\mathcal{Q}$ and show the transformations it shall go through.} $\mathbf{x}_{q} \in \mathcal{Q}$ and the list of keys $\mathcal{K}$ is then computed as “attention”:
\begin{align}
	\alpha_{q k} &\propto \exp \left(\frac{{\mathbf{x}_{q}}^{\top} W_{Q} \cdot K}{\sqrt{d}}\right) ; \forall \mathbf{x}_{k} \in \mathcal{K} \\
	\alpha_{q,:}&=\operatorname{{softmax}}\left(\frac{{\mathbf{x}_{q}}^{\top} W_{Q} \cdot K}{\sqrt{d}}\right) \in \mathbb{R}^{|\mathcal{K}|}
\end{align}
The $k$-th element $\alpha_{q k}$ in the vector $\alpha_{q,:}$ reveals the particular proximity between $\mathbf{x}_{k}$ and $\mathbf{x}_{q} .$ The computed attention values are then used as weights for the final embedding $\mathbf{x}_{q}$ :
\begin{align}
	\tilde{\mathbf{x}}_{q} &=\tau\left({\mathbf{x}_{q}}+W_{\mathbf{F C}}^{\top} \sum_{k} \alpha_{q k} V_{:, k}\right)
\end{align}
$V_{:, k}$ is the $k$-th column of $V$. $W_{\mathbf{F C}} \in \mathbb{R}^{d^{\prime} \times d}$ is the projection weights of a fully connected layer. $\tau$ completes a further transformation, which is implemented by the dropout~\cite{srivastava2014dropout} and layer normalization~\cite{ba2016layer}.

With the help of transformer, we can meta-calibrate the classifiers and prototypes into the same scale, \ie, $\tilde{W}$. Besides, the transformation process also gives the instance-specific embeddings $\tilde{\phi}(\x)$, and we can get a more proper prediction with the adapted embeddings.

\begin{figure}[t]
	\begin{center}

		{	
			\includegraphics[width=.95\columnwidth]{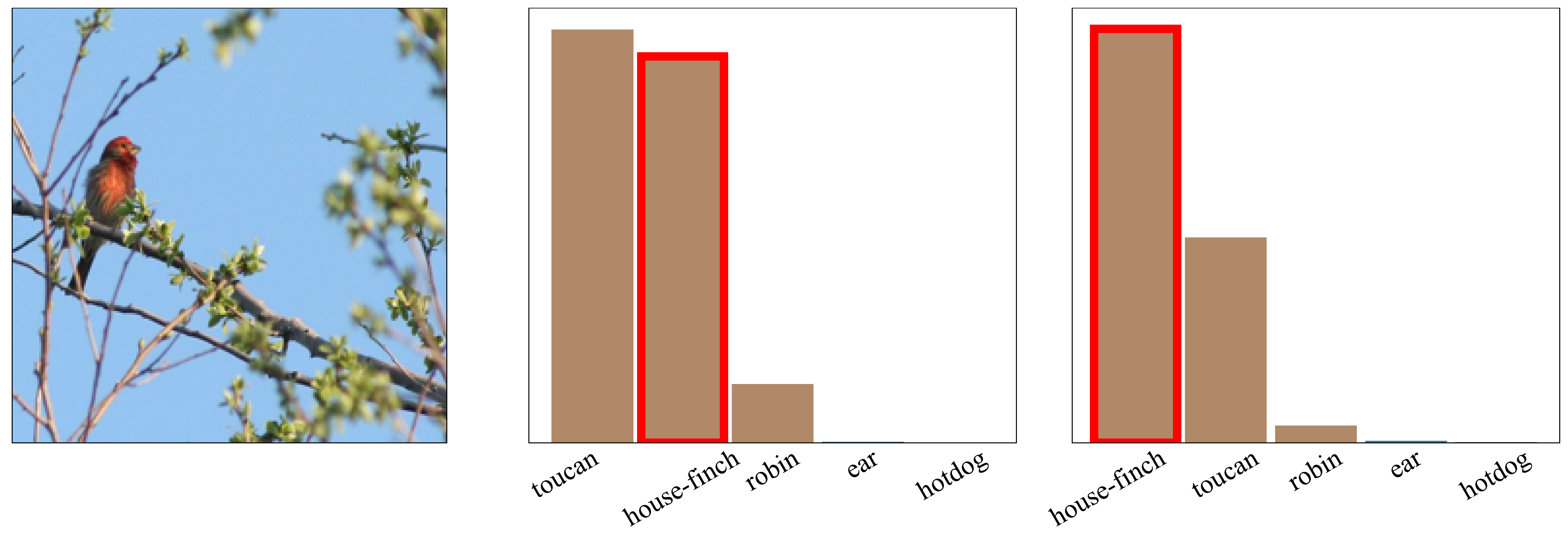}
			\includegraphics[width=.95\columnwidth]{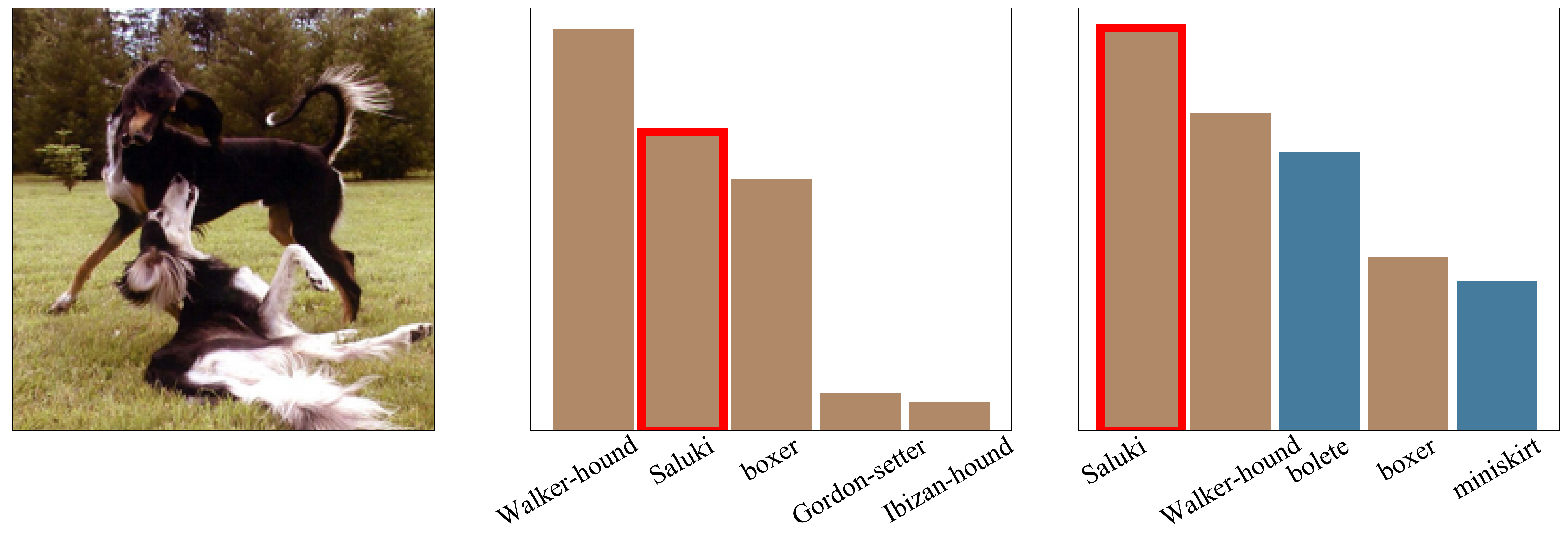}
			\includegraphics[width=.95\columnwidth]{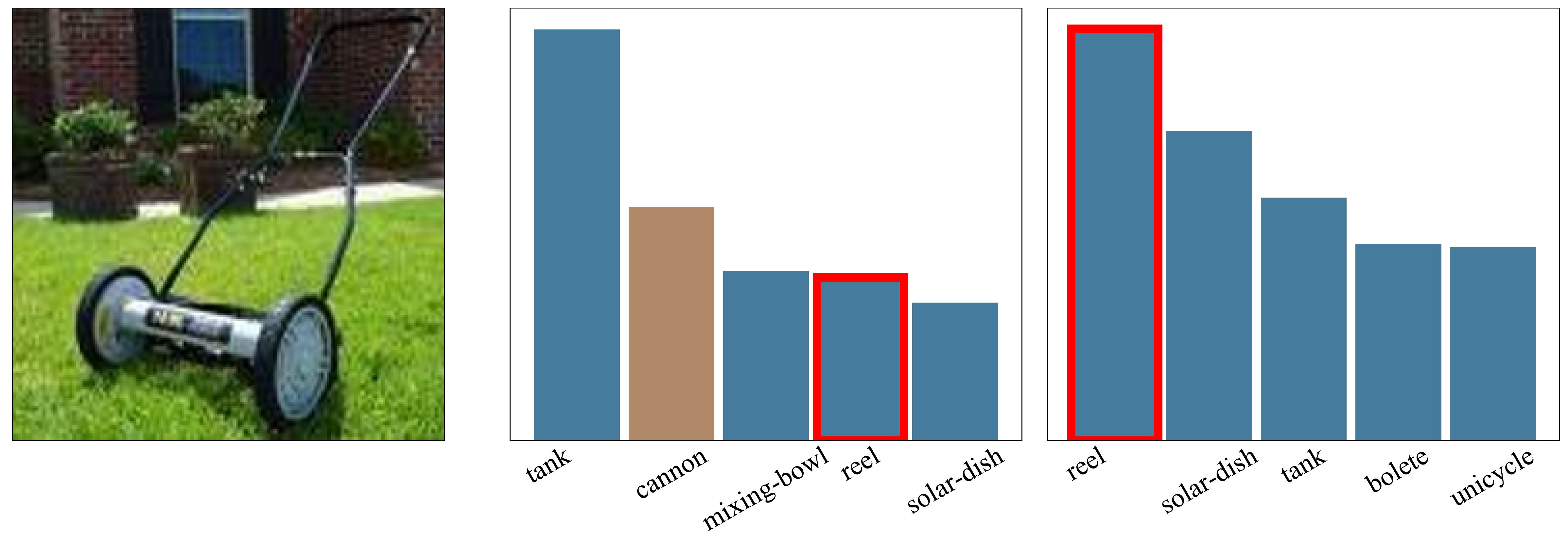}
			\includegraphics[width=.95\columnwidth]{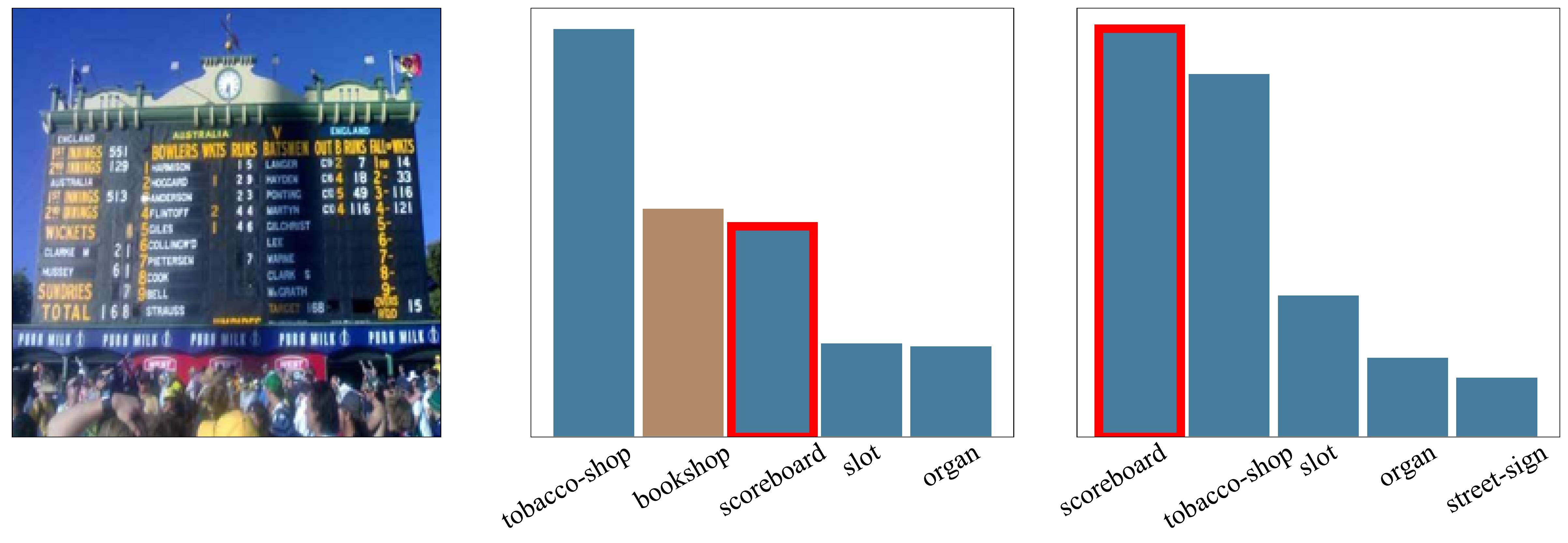}
			\includegraphics[width=.95\columnwidth]{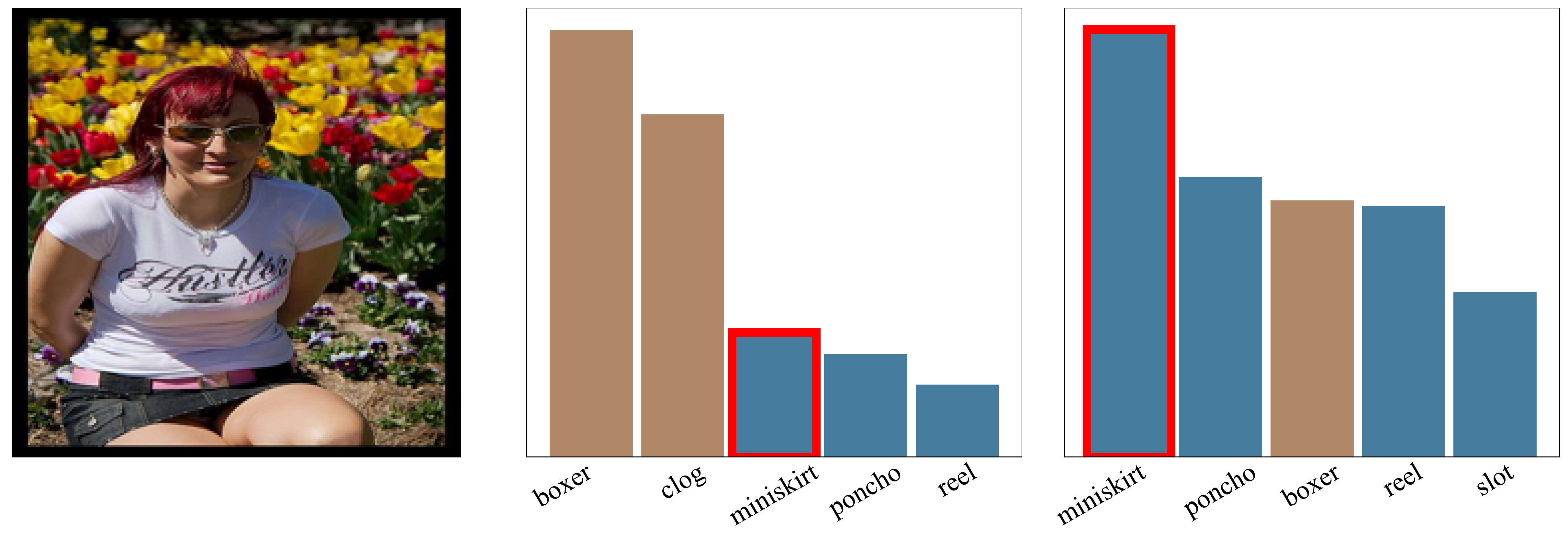}
			\includegraphics[width=.95\columnwidth]{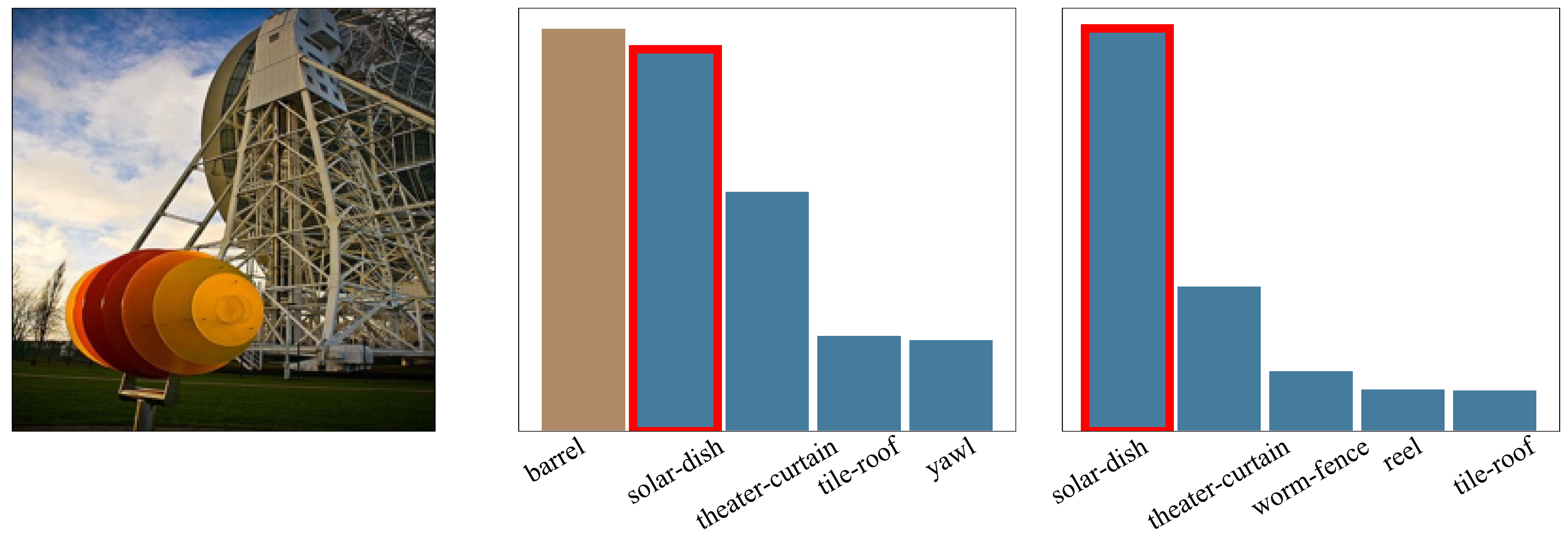}
		}
		
	\end{center}
	\caption{ Additional visualization of the prediction probability before and after meta-calibration on \textit{mini}ImageNet. The first row indicates original images. The second row indicates the top-5 output probability before meta-calibration. The third row indicates the top-5 output probability after meta-calibration.
		Base classes are shown with brown color, and incremental classes are shown in blue. The ground-truth class is shown with red edges.
	} \label{figure:meta-calibration-supp}
\end{figure}

\section{Additional Visualization Results}

In the main paper, we have shown several typical predictions before and after meta-calibration. In this section, we provide six additional images about the meta-calibration process. The results are shown in Figure~\ref{figure:meta-calibration-supp}. As we can infer from the figures, the top 2 images are from base classes, and the bottom 4 are from incremental classes. The model without contextualizing ability may misclassify a house-finch into a toucan or predict a Saluki into a Walker-hound. 
When classifying incremental classes, it shall concentrate on the irrelevant features and make incorrect predictions, \eg, predicting a reel into a tank, predicting a miniskirt into a boxer, or predicting a solar dish into a barrel.
The meta-learned calibration module is good at handling such embedding adaptation process. It aligns the embeddings with discriminative features and outputs the instance-specific embeddings. As a result, the model with meta-calibration module can get better prediction results for both base and incremental classes.

\end{document}